\documentclass[acmtog]{acmart}
\usepackage{hyperref}
\usepackage{geometry}
\AtBeginDocument{%
  }

\setcopyright{none}
\AtBeginDocument{%
  }
\setcopyright{none}
\usepackage{graphicx}
\usepackage{stfloats}
\usepackage{multirow}
\usepackage{enumitem}
\usepackage{amsmath,amsfonts,bm}
\usepackage{multirow} 

\def\eqref#1{equation~\ref{#1}}

\def\1{\bm{1}}

\DeclareMathAlphabet{\mathsfit}{\encodingdefault}{\sfdefault}{m}{sl}
\SetMathAlphabet{\mathsfit}{bold}{\encodingdefault}{\sfdefault}{bx}{n}

\definecolor{cvprblue}{rgb}{0.21,0.49,0.74}

\begin{document}

\title{ConsistEdit: Highly Consistent and Precise Training-free Visual Editing}

\author{Zixin Yin}
\orcid{0003-0443-7915}
\email{zixin.yin@connect.ust.hk}
\affiliation{%
  \institution{Hong Kong University of Science and Technology}
}

\author{Ling-Hao Chen}
\orcid{0000-0002-2528-6178}
\email{thu.lhchen@gmail.com}
\affiliation{%
  \institution{Tsinghua University}
}
\affiliation{%
  \institution{International Digital Economy Academy}
}

\author{Lionel Ni}
\orcid{0000-0002-2325-6215}
\email{ni@hkust-gz.edu.cn}
\affiliation{%
  \institution{Hong Kong University of Science and Technology, Guangzhou}
}
\affiliation{%
  \institution{Hong Kong University of Science and Technology}
}

\author{Xili Dai}
\orcid{0000-0001-5526-8934}
\email{daixili.cs@gmail.com}
\affiliation{%
  \institution{Hong Kong University of Science and Technology, Guangzhou}
}
\renewcommand{\shortauthors}{Zixin Yin, et al.}

\begin{abstract}
Recent advances in training-free attention control methods have enabled flexible and efficient text-guided editing capabilities for existing image and video generation models. However, current approaches struggle to simultaneously deliver strong editing strength while preserving consistency with the source. For instance, in color-editing tasks, they struggle to maintain structural consistency in edited regions while preserving the rest intact. This limitation becomes particularly critical in multi-round and video editing, where visual errors can accumulate over time. Moreover, most existing methods enforce global consistency, which limits their ability to modify individual attributes such as texture while preserving others, thereby hindering fine-grained editing. Recently, the architectural shift from U-Net to Multi-Modal Diffusion Transformers (MM-DiT) has brought significant improvements in generative performance and introduced a novel mechanism for integrating text and vision modalities. These advancements pave the way for overcoming challenges that previous methods failed to resolve. Through an in-depth analysis of MM-DiT, we identify three key insights into its attention mechanisms. Building on these, we propose ConsistEdit, a novel attention control method specifically tailored for MM-DiT. ConsistEdit incorporates vision-only attention control, mask-guided pre-attention fusion, and differentiated manipulation of the query, key, and value tokens to produce consistent, prompt-aligned edits. Extensive experiments demonstrate that ConsistEdit achieves state-of-the-art performance across a wide range of image and video editing tasks, including both structure-consistent and structure-inconsistent scenarios. Unlike prior methods, it is the first approach to perform editing across all inference steps and attention layers without handcraft, significantly enhancing reliability and consistency, which enables robust multi-round and multi-region editing. Furthermore, it supports progressive adjustment of structural consistency, enabling finer control. ConsistEdit represents a significant advancement in generative model editing and unlocks the full editing potential of MM-DiT architectures.  Here is the project \textcolor{cvprblue}{\href{https://zxyin.github.io/ConsistEdit}{website}}.
\end{abstract}


\begin{teaserfigure}
  \includegraphics[width=0.95\textwidth]{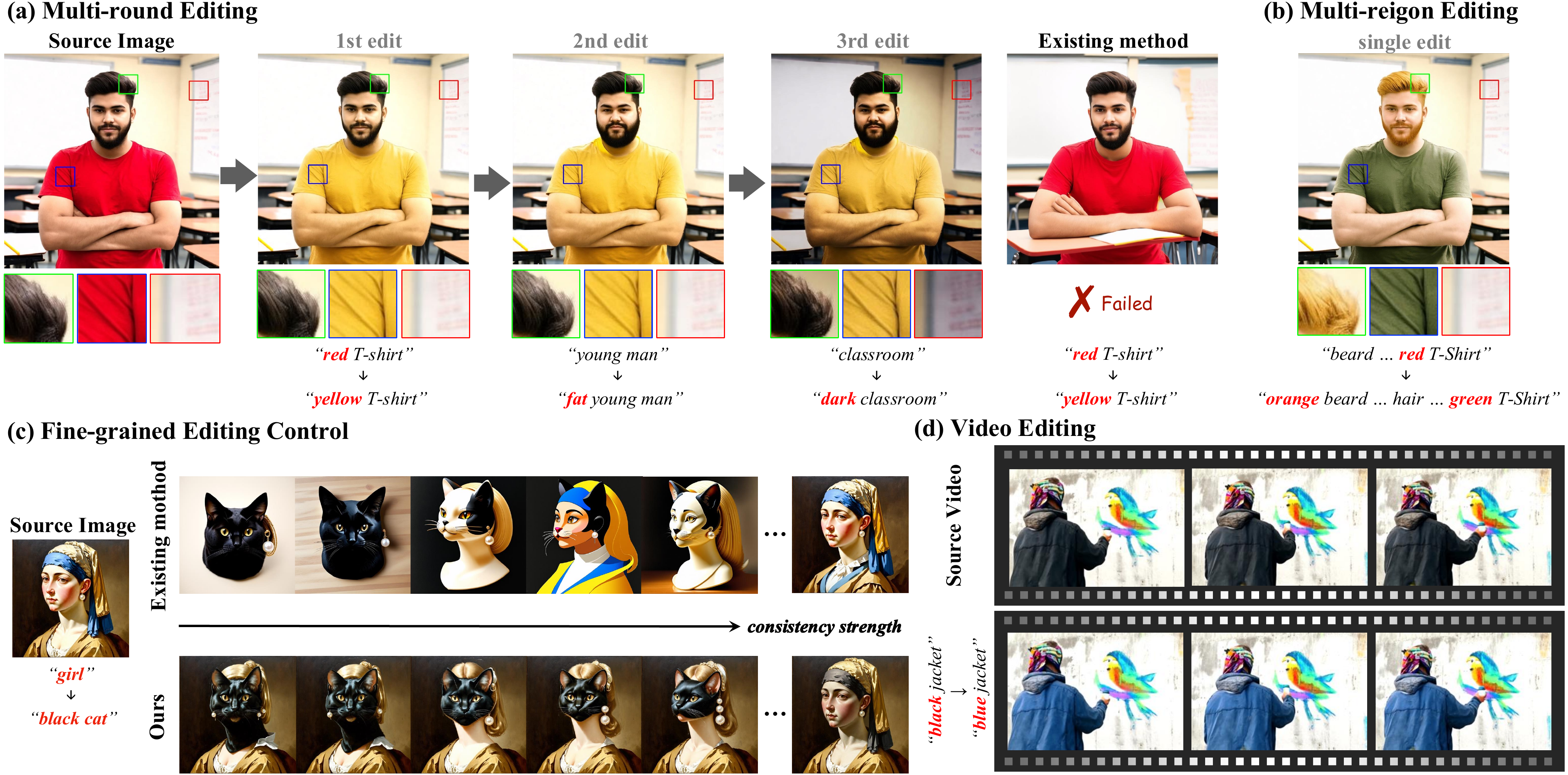}
   \caption{\textbf{(a) ConsistEdit} enables multi-round editing by allowing users to specify both the target region and the nature of the editing through prompts. Unlike existing methods, it can perform structure-preserving (hair, clothing folds) and shape-changing with identity-preserving edits in edited regions while keeping non-edited regions intact. \textbf{(b)} ConsistEdit handles multi-region edits in one pass and preserves both the edited structure and unedited content. \textbf{(c)} Our method enables smooth control over consistency strength in the edited region. In contrast, existing approaches lack smooth transitions and often alter non-edited areas. \textbf{(d)} Beyond image editing and rectified flow models, ConsistEdit generalizes well to all MM-DiT variants, including diffusion and video models.}
  \label{fig:teaser}
  \vspace{-0.5em}
\end{teaserfigure}

\maketitle

\section{Introduction}

Attention control techniques, which manipulate the query (\textit{\textbf{Q}}), key (\textit{\textbf{K}}), and value (\textit{\textbf{V}}) tokens in the attention mechanism, have been widely adopted because their training-free nature enables flexible and efficient extensions of generative models to image and video editing tasks. For example, in image editing, Prompt-to-Prompt (P2P)~\cite{hertz2022prompt} introduced a method to control cross-attention layers, enabling text-guided editing without the need for additional data or retraining. This inspired follow-up work in video editing, such as Video-P2P~\cite{liu2024video}, while Masactrl~\cite{cao2023masactrl} extended P2P from rigid to non-rigid editing. 

Despite these advancements, there are still two fundamental challenges in text-guided editing: {\color{cvprblue}{\textbf{(1)}}} the method must not only modify content in according to the input text but also preserve consistency in both editing and non-editing regions. In editing regions, for example, structure should remain stable when changing color, and the character identity must stay recognizable when adjusting shape. In non-editing regions, all visual elements should remain identical to the original. However, these two requirements are often not simultaneously satisfied in existing training-free methods~\cite{hertz2022prompt, cao2023masactrl, jiao2025uniedit}, leading to unacceptable results in tasks such as color and material modifications. As shown in part Fig.~\ref{fig:teaser}~(a), existing methods tend to introduce noticeable changes but compromise consistency in both edited and non-edited regions. Therefore, maintaining editing strength and consistency is essential for multi-round and video editing, where both iterative accumulation and additional temporal dimension can exacerbate visual errors.
{\color{cvprblue}{\textbf{(2)}}} Beyond the inability to satisfy both requirements, existing methods typically enforce overall (\textit{e.g.}, texture and structure) consistency, which severely limits performance in fine-grained editing. When a task requires preserving texture while altering structure, or vice versa, these methods often fail, see Fig.~\ref{fig:teaser}~(c). Allowing more targeted control, such as focusing consistency on structure alone, would enable more flexible and effective editing.

Amid these unresolved issues, the field of image and video generation has undergone astonishing advancements due to the transition in architectures from U-Net~\citep{rombach2022high} to Multi-Modal Diffusion Transformer~(MM-DiT)~\citep{esser2024scaling}, shedding a new light on solving these problems mentioned above. 
Since training-free attention control methods heavily depend on the underlying architecture of the generative model, it is crucial to study how to tailor it to MM-DiT. To date, only one work, DiTCtrl~\cite{cai2024ditctrl}, has investigated attention control in MM-DiT, and even that does not target editing tasks. Instead, it targeted multi-prompt long video generation. As a result, the lack of investigation into the attention mechanisms of MM-DiT in editing tasks significantly limits existing approaches~\cite{wang2024taming,deng2024fireflow}.

To address this gap, we conduct a detailed study of MM-DiT’s attention architecture, starting by contrasting it with the more commonly studied U-Net. In U-Net, cross-attention governs text guidance, while self-attention drives visual generation, resulting in a two-stage separation of modalities. In contrast, MM-DiT merges textual and visual information, applying self-attention to jointly process modalities. Through in-depth analysis and experimental exploration, we derive three key insights of MM-DiT models:
\begin{itemize}[leftmargin=*]
    \item Vision-only is crucial: Editing effectiveness relies on modifying only the vision parts, since interfering with text tokens often leads to generation instability (Fig.~\ref{fig:attn_fusion}).
    \item Homogeneous for all layers: Visualizations of the vision parts of \textbf{\textit{Q}}, \textbf{\textit{K}}, and \textbf{\textit{V}} across attention layers (Fig.~\ref{fig:attn_visualize}) show that, unlike U-Net, each layer in MM-DiT retains rich semantic content. Thus, attention control must be applied to all layers.
    \item Strong structure controllability from \textbf{\textit{Q}} and \textbf{\textit{K}}: Applying attention control solely on the vision parts of \textbf{\textit{Q}}, \textbf{\textit{K}} results in strong controllable structural preservation (Fig.~\ref{fig:disentangle}).
\end{itemize}
By grounding these insights, we introduce ConsistEdit, a novel attention control method specifically adapted to MM-DiT to address the challenges through three core operations: (1) Vision-only attention control: attention control is applied solely to the vision parts across all layers; (2) Pre-attention mask fusion: editing and non-editing regions are fused before the attention calculation; (3) Differentiated control for \textit{\textbf{Q}}, \textit{\textbf{K}}, and \textit{\textbf{V}}: we apply distinct control mechanisms to \textit{\textbf{Q}} and \textit{\textbf{K}} for structure, and \textit{\textbf{V}} for content.

Through extensive experiments, we show that ConsistEdit enables structurally consistent at finer levels such as lighting and shading in edited regions, while keeping non-edited regions unchanged.
As a result, ConsistEdit can address the two fundamental challenges in text-guided editing mentioned before: {\color{cvprblue}{\textbf{(1)}}} ConsistEdit achieves \textbf{state-of-the-art (SOTA)} performance across diverse editing tasks including structure-consistency and -inconsistency tasks, enabling iterative multi-round editing, as well as single-pass multi-region editing, see Fig.~\ref{fig:teaser}~(a)~(b). Additionally, it demonstrates strong generalization across diverse generation models and editing tasks, including video editing, showcasing its versatility and practical potential, as shown in Fig.~\ref{fig:teaser}~(d) and~\ref{fig:flux}.  {\color{cvprblue}{\textbf{(2)}}} Instead of enforcing overall consistency, ConsistEdit supports progressive adjustment of structural consistency, allowing fine-grained control in various downstream tasks, as shown in Fig.~\ref{fig:teaser}~(c). 

To our best knowledge, ConsistEdit is the \textbf{first} approach that enables editing across all steps and layers without manual parameter adjustment, significantly improving reliability and consistency. Overall, we list our contributions as follows.

\begin{itemize}[leftmargin=*]
    \item We identify three key insights from MM-DiT foundation generation models that enable effective training-free attention control for editing tasks.
    \item We propose ConsistEdit, a novel attention control approach designed to extend the editing capabilities of MM-DiT-based models.
    \item Our method supports both structure-consistent and -inconsistent edits while maintaining fidelity in non-edited regions. Extensive experiments demonstrate that ConsistEdit sets new SOTA results in both image and video editing tasks, and enables reliable multi-round and multi-region editing.
\end{itemize}

\section{Related Work}

\subsection{Text-to-image/video Generation}
Early visual generation methods were primarily based on GANs~\cite{tao2022df,reed2016generative,yu2023talking,wang2023progressive} due to their ability to synthesize high-fidelity content. However, diffusion models~\cite{reed2016generative,ho2020denoising,rombach2022high,guo2023animatediff} have demonstrated superior generative performance, with U-Net-based architectures~\cite{rombach2022high} becoming the dominant paradigm. As U-Net designs encounter scalability limitations in data and model parameter size, the field has progressively shifted toward transformer-based architectures, particularly diffusion transformers (DiT)~\cite{peebles2023scalable}. Among these, MM-DiT~\cite{esser2024scaling} has emerged as a widely adopted backbone for text-conditional generation. Many recent state-of-the-art models~\cite{esser2024scaling,sd3.5_2024,flux2024,yang2024cogvideox,kong2024hunyuanvideo,liu2024generative} build upon MM-DiT, achieving remarkable performance, such as SD3~\cite{esser2024scaling} and FLUX~\cite{flux2024} for image generation, as well as CogVideoX~\cite{yang2024cogvideox} for video generation. In this work, we tailor a new attention control method for MM-DiT-based models.
\vspace{-1em}
\subsection{Text-guided Editing} Early work focused on training-based approaches that leveraged generative models to achieve controllable image editing~\cite{karras2019style}. With the rapid advancement of generative models, attention has shifted toward training-free editing methods, which offer greater flexibility and efficiency. These training-free approaches generally fall into two categories: sampling-based and attention-based methods. Sampling-based approaches introduce controlled randomness into the generation process to enable flexible editing~\cite{jiao2025uniedit,huberman2024edit,kulikov2024flowedit}, while attention-based methods achieve editing ability by directly altering attention tokens. Prompt-to-Prompt~\cite{hertz2022prompt} was the first to introduce attention manipulation on the cross-attention layers of U-Net, adopted in many subsequent editing methods~\cite{chen2024training,yang2023dynamic}. Video-P2P~\cite{liu2024video} extends this cross-attention control to video editing. FateZero~\cite{qi2023fatezero} combines self-attention with a blending mask derived from cross-attention features of the source prompt. Methods such as MasaCtrl~\cite{cao2023masactrl} and DiTCtrl~\cite{cai2024ditctrl} employ similar attention control strategies, applied to U-Net and MM-DiT architectures respectively. Despite their differences, all existing attention control methods can be understood as multi-branch frameworks~\cite{cao2023masactrl,cai2024ditctrl,ju2023direct,wang2024taming,rout2024semantic}, and can be uniformly expressed as special cases of Eq.~\ref{eq:attention_tg}. Notably, all above methods rely on selectively manipulating specific inference steps or attention layers, which limits their robustness and consistency with respect to the source. In contrast, our approach is the first one requires no manual selection of steps or layers.

\section{Method}

\subsection{Preliminary}

\subsubsection{Generation procedure.} The current visual generation procedure is a systematical method which includes generation algorithm and foundation network architecture. 
$\bm{z}^{T}\xrightarrow{}\bm{z}^{T-1}\xrightarrow{}\dots\xrightarrow{}\bm{z}^{t}\xrightarrow{}\dots\xrightarrow{}\bm{z}^{0}$ shows the procedure for generating the final image or video from random noise $\bm{z}^{T}$ in $T$ steps. The generation algorithm could be latent diffusion, flow matching, or rectified flow.

Beyond the generation algorithm, the foundation network architecture plays a crucial role in affecting the final generation results. In each step, the network $f(\cdot)$ integrates the text prompt tokens $\bm{\mathrm{P}}$ and the result of previous step $\bm{z}^t$ to generate the result of the next step $\bm{z}^{t-1}$: $ \bm{z}^{t}\xrightarrow{\hspace{2mm} f(\bm{z}^{t}|\bm{\mathrm{P}})\hspace{2mm}}\bm{z}^{t-1}$.
The network $f(\cdot)$ can be U-Net or MM-DiT. It takes the pair of $(\bm{z}^{t},\bm{\mathrm{P}})$ as input, which present the vision $\bm{z}^{t}$ and text $\bm{\mathrm{P}}$ tokens respectively. It goes through each layer of the network, and Eq.~\ref{eq:attention} shows how each attention layer works.
\begin{equation}
\begin{gathered}
  \{\bm{z}^{t}(l), \bm{\mathrm{P}}(l)\} 
    \xrightarrow{\hspace{2mm}\bm{g}(\cdot)\hspace{2mm}} \bm{Q}^{l}, \bm{K}^{l}, \bm{V}^{l}, \\
  \bm{z}^{t}(l+1) 
    = \mathrm{Attention}(\bm{Q}^{l}, \bm{K}^{l}, \bm{V}^{l}) 
    = \bm{V}^{l} \cdot \mathrm{Softmax}\left(\frac{\bm{Q}^{l}(\bm{K}^{l})^{\top}}{\sqrt{d}}\right).
\end{gathered}
\label{eq:attention}
\end{equation}
\noindent We unify the formulation of cross-attention and self-attention in Eq.~\ref{eq:attention}. The function $\bm{g}(\cdot)$ denotes a pre-attention operation that plays different roles in cross-attention and self-attention layers. Specifically, in the $l$-th layer of a U-Net, if it is a cross-attention layer, $\bm{g}(\cdot)$ maps the text tokens $\bm{\mathrm{P}}(l)$ to $\bm{K}^l$ and $\bm{V}^l$, and maps the vision tokens $\bm{z}^{t}(l)$ to $\bm{Q}^l$. In contrast, for self-attention layers, $\bm{g}(\cdot)$ ignores the text tokens and maps $\bm{z}^{t}(l)$ to all of $\bm{Q}^l$, $\bm{K}^l$, and $\bm{V}^l$.

In contrast, MM-DiT is a self-attention-only architecture, without cross-attention. 
Before computing attention, each MM-DiT block applies a pre-attention transformation $\bm{g}(\cdot)$, which includes operations such as MLP modulation, residual connections, normalization, and other components.
In each block, the pre-attention $\bm{g}(\cdot)$ maps the vision $\bm{z}^{t}(l)$ and text $\bm{\mathrm{P}}(l)$ tokens respectively and concatenate every vision and text pair to get $\bm{Q}^l$, $\bm{K}^l$, $\bm{V}^l$. 
In other words, $\bm{Q}^l$, $\bm{K}^l$, $\bm{V}^l$ all contain text and vision parts.

\subsubsection{Inversion.} The inversion procedure aims to accurately reverse the generation process to recover the initial noise $\bm{z}^{T}$ that can reconstruct the real image or video tokens $\bm{z}^{0}$.

\subsubsection{Editing.} The original editing method can trace back to the image processing era~\citep{jahne2005digital} and the task was formulated as follows:
\begin{equation}
  \bm{I}_{tg}=(\bm{1}-\bm{M})\odot\bm{I}_{s}+\bm{M}\odot\bm{I}_{e},
  \label{eq:editing}
\end{equation}
\noindent where the user offers source image $\bm{I}_s$ and editing regions (mask $\bm{M}$). The goal of the editing task was to generate the edited content $\bm{I}_e$ and then blend it back to the source image while preserving the non-edited regions of the source image. $\odot$ denotes the element-wise Hadamard product of two matrices. 

\subsubsection{Attention control approach for training-free editing.} The current visual editing methods in the background of generation models~\cite{hertz2022prompt,cao2023masactrl}, leverage the attention control approach to extend the editing capability of the foundation generation model in a training-free manner. In concrete, they employ a dual-network architecture: one network is dedicated to reconstructing the original source given the prompt tokens $\bm{\mathrm{P}}_s$ and random noise $\bm{z}^T$, while the other is focused on editing. The dual-network shares the same network parameters. 

We formulate the procedure of the editing in a way of dual-network architecture. By applying the generation process to the source $\bm{I}_s$, we obtain the full generation trajectory of the source tokens:
$\bm{z}^{T} \rightarrow \bm{z}^{T-1}_s \rightarrow \dots \rightarrow \bm{z}^{t}_s \rightarrow \dots \rightarrow \bm{z}^{0}_s$.
At each step, the update follows $\bm{z}^{t}_s \xrightarrow{\hspace{2mm} f(\bm{z}^{t}_s|\bm{\mathrm{P}}_s) \hspace{2mm}} \bm{z}^{t-1}_s$,
and each attention layer is computed as $\bm{z}^{t}_s(l+1) = \mathrm{Attention}(\bm{Q}^{l}_s, \bm{K}^{l}_s, \bm{V}^{l}_s)$.

The generation procedure for the target $\bm{I}_{tg}$ starting from the same noise, $\bm{z}^{T}\rightarrow\bm{z}^{T-1}_{tg}\rightarrow\dots\rightarrow\bm{z}^{t}_{tg}\rightarrow\dots\rightarrow\bm{z}^{0}_{tg}$, and each step $\bm{z}^{t}_{tg}\xrightarrow{\hspace{2mm} f(\bm{z}^{t}_{tg}|\bm{\mathrm{P}}_t)\hspace{2mm}}\bm{z}^{t-1}_{tg}$, are very similar to that of the source, but with different attention operation which we call it attention control.

When we get the $\bm{Q}^{l}_{tg},\bm{K}^{l}_{tg},\bm{V}^{l}_{tg}$ from the $l$-th attention layer of $t$-th step in the target generation procedure, the attention control no longer apply them directly to attention operation, but replace some of them from the generation procedure of the source.
\begin{equation}
\begin{gathered}
  \{\bm{z}^{t}_{tg}(l),\bm{\mathrm{P}}_{tg}\}\xrightarrow{\hspace{2mm}\bm{g}(\cdot)\hspace{2mm}}\bm{Q}^{l}_{tg},\bm{K}^{l}_{tg},\bm{V}^{l}_{tg},\\
  f^{t}_o=\mathrm{Attention}(\bm{Q}^{l}_{tg},{\color{cvprblue}\bm{K}^{l}_{s},\bm{V}^{l}_{s}}|\bm{M}_c),\\
  f^{t}_b=\mathrm{Attention}(\bm{Q}^{l}_{tg},{\color{cvprblue}\bm{K}^{l}_{s},\bm{V}^{l}_{s}}|\bm{1}-\bm{M}_c),\\
  \bm{z}^{t}_{tg}(l+1)=(\bm{1}-\bm{M})\odot f^{t}_b+\bm{M}\odot f^{t}_o.
\end{gathered}
\label{eq:attention_tg}
\end{equation}
Eq.~\ref{eq:attention_tg} is a example of attention control formulation of MasaCtrl~\cite{cao2023masactrl} and DiTCtrl~\cite{cai2024ditctrl}, where they replace the $\bm{K},\bm{V}$ from the source in self-attention layers. Here, $\bm{M}_c$ and $\bm{M}$ donate masks extracted from attention maps\footnote{DiTCtrl adopts all-one masks for both $\bm{M}_c$ and $\bm{M}$ in editing tasks, despite using extracted masks for long video generation.} of the source and target generation procedure, respectively. $\bm{M}$ is used as the attention mask. 
\subsection{ConsistEdit: A New Attention Control for MM-DiT}
\subsubsection{The analysis of the attention mechanism in MM-DiT}
MM-DiT~\cite{esser2024scaling} fundamentally differs from U-Net~\cite{rombach2022high} in its attention mechanism. In U-Net, cross-attention handles text guidance, while self-attention focuses on visual content generation, creating a two-stage process. In contrast, MM-DiT merges text and visual information and employs self-attention to process both modalities simultaneously. This architecture shift renders existing U-Net-based attention control methods ineffective.
\begin{figure}[!t]
  \centering
  \includegraphics[width=\linewidth]{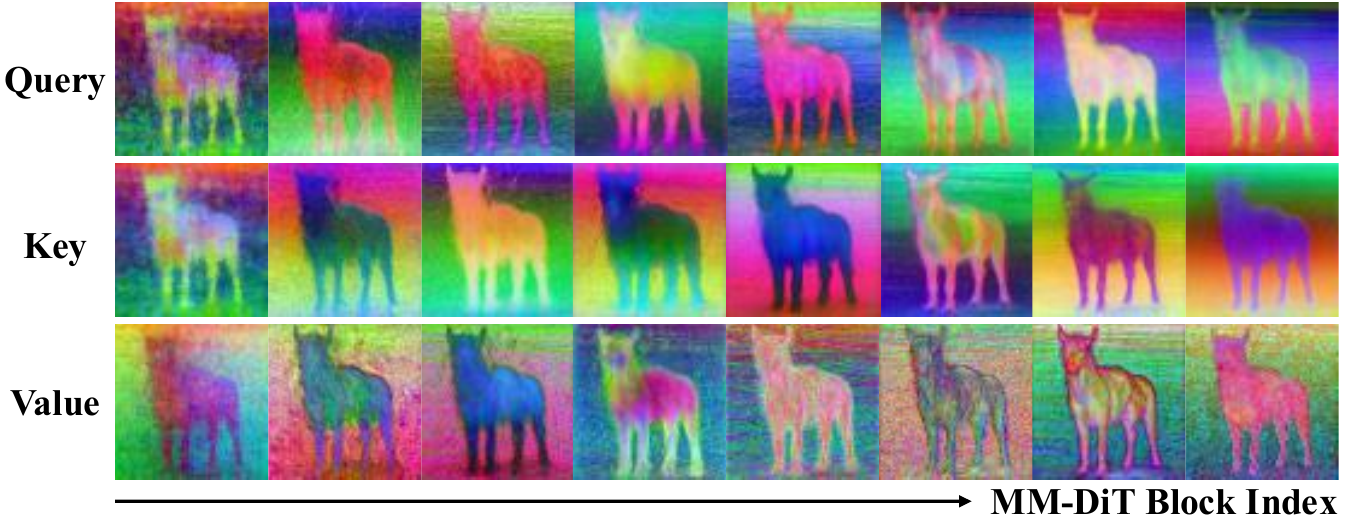}
  \caption{ Visualization of projected \textit{\textbf{Q}}, \textbf{\textit{K}}, \textbf{\textit{V}} vision tokens in attention layers of the MM-DiT blocks at 15th sampling step of prompt ``A standing horse.'' }
  \label{fig:attn_visualize}
\end{figure}

\begin{figure*}[t!]
\centering
\includegraphics[width=0.95\textwidth]{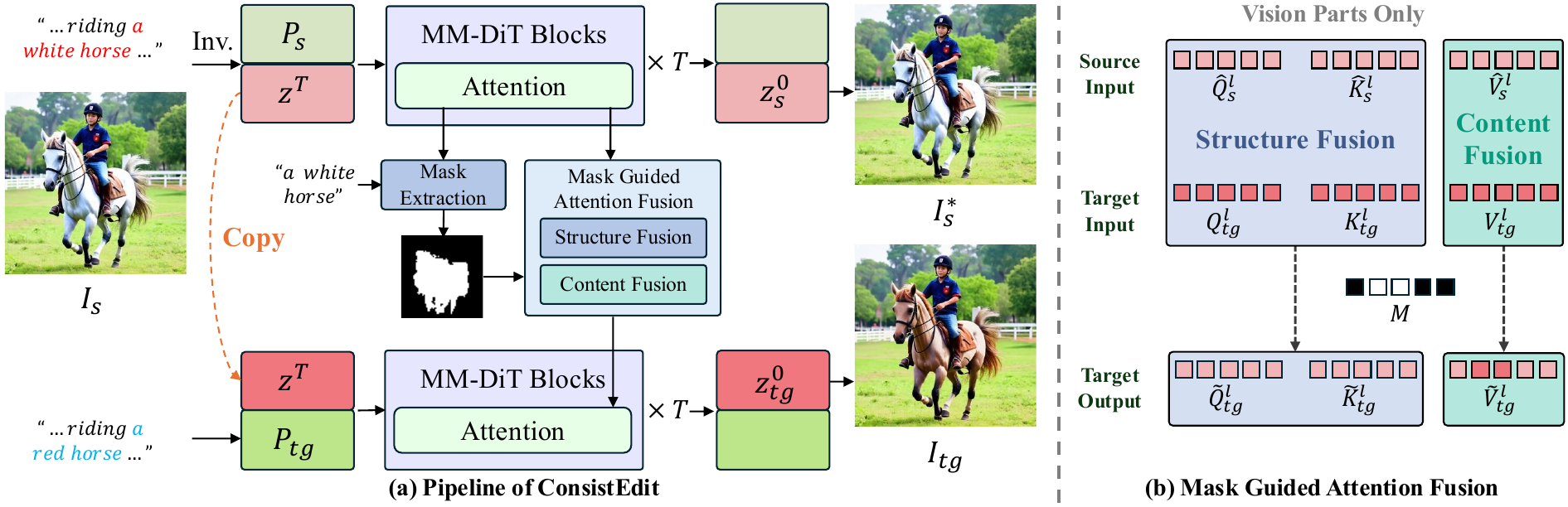}
\caption{ \textbf{(a)} shows the \textbf{ConsistEdit} pipeline. Given a real image or video $\bm{I}_s$ and source text tokens $\bm{\mathrm{P}}_s$, we first invert the source to obtain the vision tokens $\bm{z}^T$, which is concatenated with the target prompt tokens $\bm{\mathrm{P}}_{tg}$ and passed into the generation process to produce the edited image or video $\bm{I}_{tg}$. During inference, a mask $\bm{M}$ generated by our extraction method delineates editing and non-editing regions. We apply structure and content fusion to enable prompt-aligned edits while preserving structural consistency within edited regions and maintaining content integrity elsewhere. \textbf{(b)} illustrates the mask-guided attention fusion for timesteps where $t>(1-\alpha)T$, which is applied exclusively to the vision parts, while the text parts remain unchanged.}
\label{fig:pipeline}
\end{figure*}
\begin{figure}[t]
  \centering
  \includegraphics[width=\linewidth]{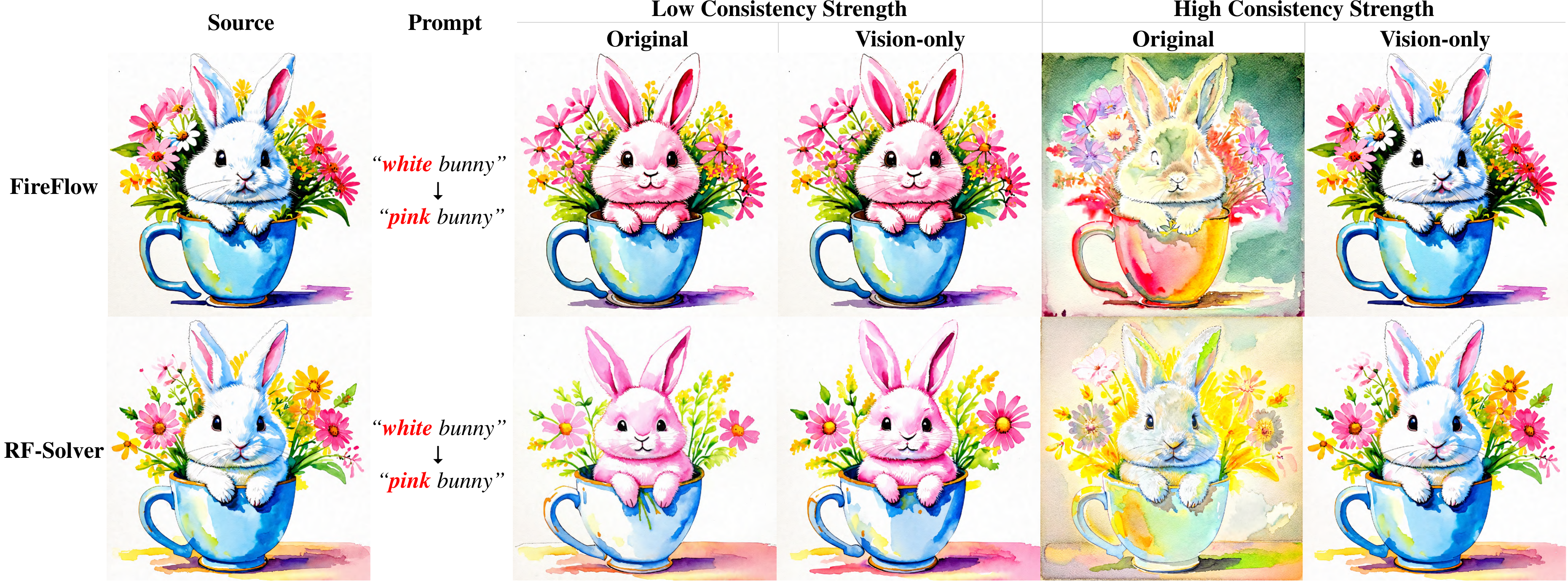}
  \caption{Comparison of \textbf{\textit{V}} token swapping strategies for content consistency.
Swapping vision-only \textbf{\textit{V}} tokens leads to superior content consistency under high consistency strength settings, while maintaining comparable editing capability to original methods when the consistency strength is low.}
  \label{fig:attn_fusion}
\end{figure}
For instance, DiTCtrl~\cite{cai2024ditctrl} adopts the strategy of MasaCtrl~\cite{cao2023masactrl} by applying attention control to the latter blocks of the model. This design originates from the downsampling–upsampling structure of the U-Net encoder–decoder architecture, where MasaCtrl performs edits in the decoder stages. However, MM-DiT does not exhibit such stage separation, as it lacks a distinct decoder stage on which editing can be focused, as illustrated by the PCA decomposition visualization in Fig.~\ref{fig:attn_visualize}. Consequently, directly transferring this strategy leads to structural artifacts, as shown in Fig.~\ref{fig:edit_strength_compare}, Fig.~\ref{fig:consistent_compare}, and Fig.~\ref{fig:inconsistent_compare}. Further experiments (Fig.~\ref{fig:ablation_swap_layer}) confirm that editing across all blocks yields superior results.

Moreover, Fig.~\ref{fig:attn_fusion} compares FireFlow~\cite{deng2024fireflow} and RF-Solver~\cite{wang2024taming}, each using their original attention control method on either all parts (original) or vision-only parts of tokens. Under higher consistency strength, the original approach often fails, while vision-only edits better preserve the source content. At lower consistency strength, both approaches perform similarly. These results clearly \textit{highlight} that restricting attention control to the vision parts is critical for robust editing. The detailed implementations are provided in Appendix~\ref{sec:imp_details}.

Hence, we would execute the attention control on the vision parts across all blocks. ${\color{cvprblue}\bm{\hat{Q}}^{l}_{s}}$, ${\color{cvprblue}\bm{\hat{K}}^{l}_{s}}$ and ${\color{cvprblue}\bm{\hat{V}}^{l}_{s}}$ have the same vision parts as $\bm{Q}^{l}_{s}$, $\bm{K}^{l}_{s}$ and $\bm{V}^{l}_{s}$ but the text parts are from $\bm{Q}^{l}_{tg}$, $\bm{K}^{l}_{tg}$ and $\bm{V}^{l}_{tg}$.

\subsubsection{Structural consistency in edited region.}
Besides the visual parts only replaced from source components, we also move the blending procedure before the attention operation. Then, after extensive exploration of all potential combinations of $\bm{Q}$, $\bm{K}$, and $\bm{V}$ from the source and target generation procedure, we find the combination shown in Eq.~\ref{eq:our_attention_qk} best preserves structural consistency. The mask $\bm{M}$, representing the editing region, is extracted from the source attention maps similar to~\citet{cai2024ditctrl} and is applied only to the vision parts. We refer to this method as \textit{Structure Fusion}. Furthermore, the spatially-resolved visualizations in Fig.~\ref{fig:attn_visualize} enable us to perform mask blending directly based on spatial regions. To enable controllable editing strength, we define the \textbf{consistency strength} $\alpha$ as a ratio of steps for applying attention control, which determines the level of structural preservation during editing. Technically, in structure consist editing, the attention calculation can be formulated as:
\begin{equation}
\begin{gathered}
 \bm{\tilde{Q}}^{l}_{tg} = 
  \begin{cases}
    \bm{M} \odot {\color{cvprblue}\bm{\hat{Q}}^{l}_{s}} + (\bm{1} - \bm{M}) \odot \bm{Q}^{l}_{tg}, & \text{if } t > (1-\alpha)T \\
    \bm{M} \odot \bm{Q}^{l}_{tg} + (\bm{1} - \bm{M}) \odot \bm{Q}^{l}_{tg}, & \text{otherwise}
  \end{cases},\\
  \bm{\tilde{K}}^{l}_{tg} = 
  \begin{cases}
    \bm{M} \odot {\color{cvprblue}\bm{\hat{K}}^{l}_{s}} + (\bm{1} - \bm{M}) \odot \bm{K}^{l}_{tg}, & \text{if } t > (1-\alpha)T \\
    \bm{M} \odot \bm{K}^{l}_{tg} + (\bm{1} - \bm{M}) \odot \bm{K}^{l}_{tg}, & \text{otherwise}
  \end{cases},\\
  \bm{z}^t_{tg}(l+1)=\mathrm{Attention}(\bm{\tilde{Q}}^{l}_{tg},\bm{\tilde{K}}^{l}_{tg},\bm{V}^{l}_{tg}).
\end{gathered}
\label{eq:our_attention_qk}
\end{equation}
This operation enforces structural consistency while enabling precise text control to adjust appearance and texture.

\subsubsection{Content preservation in non-edited region.}
We find that using $\bm{\hat{Q}}^{l}_{s}$ and $\bm{\hat{K}}^{l}_{s}$ in the non-editing regions can maintain structural consistency, but often leads to color shifts. To achieve high-fidelity content preservation, we further use $\bm{\hat{V}}^{l}_{s}$ in non-editing regions, which yields the best results. We refer to following strategy as \textit{Content Fusion}.

As a result, Eq.~\ref{eq:our_attention_qkv_all} defines the final formulation of ConsistEdit:
\begin{equation}
\begin{gathered}
 \bm{\tilde{Q}}^{l}_{tg} = 
  \begin{cases}
    \bm{M} \odot {\color{cvprblue}\bm{\hat{Q}}^{l}_{s}} + (\bm{1} - \bm{M}) \odot \color{cvprblue}\bm{\hat{Q}}^{l}_{s}, & \text{if } t > (1-\alpha)T \\
    \bm{M} \odot \bm{Q}^{l}_{tg} + (\bm{1} - \bm{M}) \odot \color{cvprblue}\bm{\hat{Q}}^{l}_{s}, & \text{otherwise}
  \end{cases},\\
  \bm{\tilde{K}}^{l}_{tg} = 
  \begin{cases}
    \bm{M} \odot {\color{cvprblue}\bm{\hat{K}}^{l}_{s}} + (\bm{1} - \bm{M}) \odot \color{cvprblue}\bm{\hat{K}}^{l}_{s}, & \text{if } t > (1-\alpha)T \\
    \bm{M} \odot \bm{K}^{l}_{tg} + (\bm{1} - \bm{M}) \odot \color{cvprblue}\bm{\hat{K}}^{l}_{s}, & \text{otherwise}
  \end{cases},\\
  \bm{\tilde{V}}^{l}_{tg}=\bm{M}\odot \bm{V}^{l}_{tg}+(\bm{1}-\bm{M})\odot{\color{cvprblue}\bm{\hat{V}}^{l}_{s}},\\
  \bm{z}^t_{tg}(l+1)=\mathrm{Attention}(\bm{\tilde{Q}}^{l}_{tg},\bm{\tilde{K}}^{l}_{tg},\bm{\tilde{V}}^{l}_{tg}).
\end{gathered}
\label{eq:our_attention_qkv_all}
\end{equation}

\section{Experiments}

\subsection{Setup}
\subsubsection{Baselines.} We compare our method against several recent state-of-the-art approaches built upon MM-DiT, including UniEdit-Flow~\cite{jiao2025uniedit}, DiTCtrl~\cite{cai2024ditctrl}, FireFlow~\cite{deng2024fireflow}, RF-Solver~\cite{wang2024taming}, and SDEdit~\cite{meng2021sdedit}. We focus exclusively on MM-DiT-based baselines, as previous works~\cite{deng2024fireflow, jiao2025uniedit} and our preliminary experiments (Fig.~\ref{fig:hat_compare}) show that U-Net-based methods perfom significantly worse. Methods (SDEdit) that can be adapted to MM-DiT are included in comparisons, while those cannot be transferred are excluded.

\subsubsection{Implementation.} We primarily conduct experiments using Stable Diffusion 3 Medium (\textit{a.k.a.} SD3)~\cite{esser2024scaling} for image generation and CogVideoX-2B~\cite{yang2024cogvideox} for video generation, both of which employ a pure MM-DiT architecture. Unless otherwise specified, we use the Euler sampler and adopt UniEdit-Flow~\cite{jiao2025uniedit} for inversion. For all baseline methods, we carefully tune the hyperparameters to ensure a fair comparison. Implementation details are provided in Appendix~\ref{sec:imp_details}.

\subsubsection{Benchmark.} We adopt prompts from PIE-Bench~\cite{ju2023direct} which comprises 700 editing pairs across 10 types of edits. Although our method is fully compatible with inversion methods, we adopt a noise-to-image setting to better isolate and highlight the editing capabilities, minimizing the influence of reconstruction and inversion quality. To ensure fair comparison across baselines, we use a fixed sampler and identical random seeds for each method within a comparison group, so that source images are consistent across all methods. For structure-consistent image editing, we adopt prompts on two tasks that require preserving the original structure: \textit{change color} and \textit{change material}, covering 80 image pairs in total. For structure-inconsistency image editing, we use the remaining cateogries including \textit{Change Object}, \textit{Add Object}, \textit{Delete Object}, \textit{Change Content}, \textit{Change Style}, etc.\

\subsubsection{Metrics and settings.}
Unlike the original PIE-Bench, which uses structural distance~\cite{tumanyan2022splicing} to evaluate structural similarity, we employ the Structural Similarity Index (SSIM)~\cite{wang2004image} computed on Canny edge maps~\cite{canny1986computational} borrowed from ~\citet{zhao2023uni} for a more accurate assessment. To evaluate the preservation of non-edited regions (\textit{a.k.a.} BG preservation), we compute PSNR and SSIM exclusively on those regions, which are manually annotated by human annotators. The semantic alignment of the edits is assessed using CLIP similarity~\cite{radford2021learning}, applied to both the entire image and the edited regions. 
\subsection{Quantitative Evaluation}
While prior editing methods~\cite{hertz2022prompt,cao2023masactrl,cai2024ditctrl} typically lack quantitative evaluation, we incorporate evaluation metrics inspired by related tasks (\textit{i.e.} PIE-Bench) to more effectively showcase the capabilities of our method.
\begin{figure*}[t!]
\centering
\includegraphics[width=0.9\textwidth]{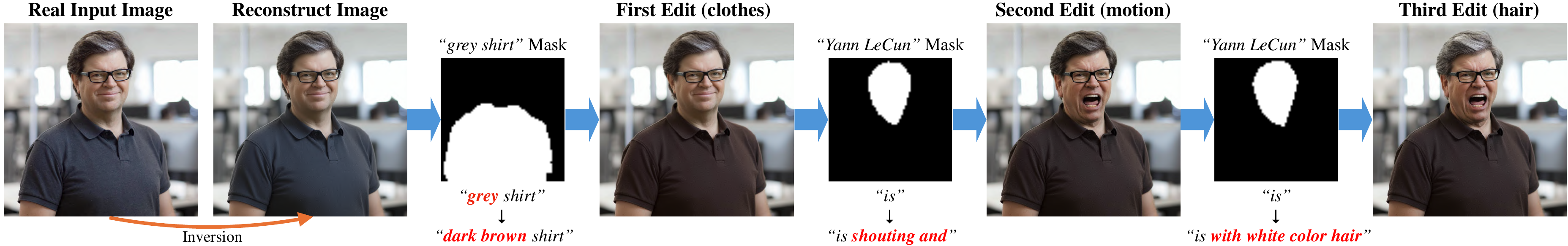}
\caption{ Real image multi-round editing results. Starting from a real image, we first perform inversion to project it into the latent space. We then sequentially edit the clothing color, motion, and hair. }
\label{fig:real_edit}
\end{figure*}
\begin{figure}[t]
  \centering
  \includegraphics[width=\linewidth]{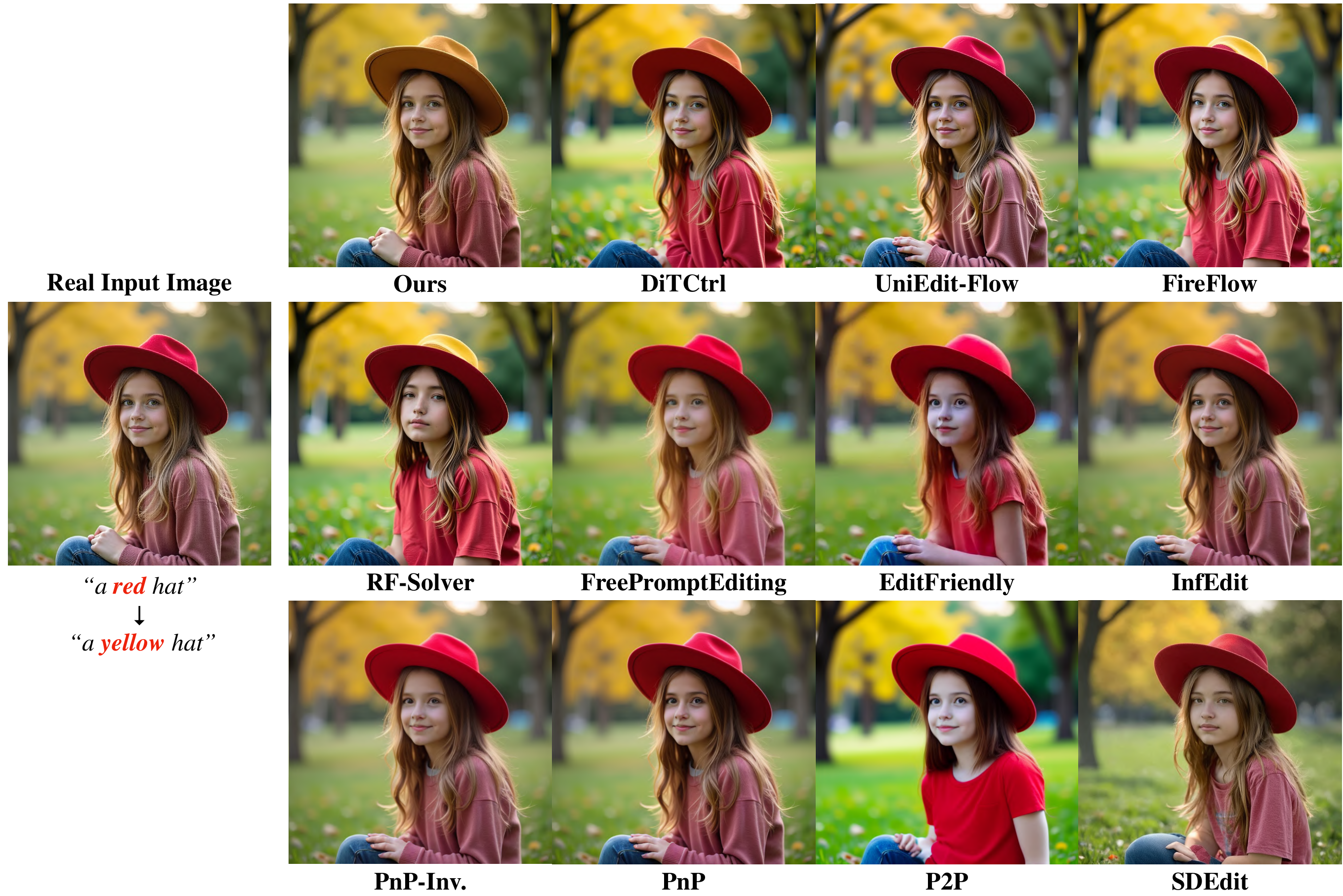}
  \caption{Qualitative comparison of methods on real image editing tasks. }
  \label{fig:hat_compare}
\end{figure}
\begin{table}[t]
\small
\centering
\caption{ Quantitative results of structural consistency comparison with RF-Solver and FireFlow using Canny SSIM~$\uparrow$.}
\begin{tabular}{lcc}
\toprule
\multicolumn{1}{c}{\multirow{2}{*}{Edit Method}} & 
\multicolumn{2}{c}{{Sampler}} \\
\cmidrule(lr){2-3}
& {RF-Solver} & {FireFlow} \\
\midrule
Fix seed & 0.5507 & 0.5557 \\
RF-Solver~\cite{wang2024taming} & 0.6225 & --- \\
FireFlow~\cite{deng2024fireflow} & --- & 0.5136 \\
Ours & \textbf{0.8714} & \textbf{0.8776} \\
\bottomrule
\end{tabular}
\label{table:canny-ssim-compare}
\end{table}

\subsubsection{Evaluation results.}
Tab.~\ref{table:canny-ssim-compare} reports a structural consistency comparison with RF-Solver~\cite{wang2024taming} and FireFlow~\cite{deng2024fireflow}. We evaluate each baseline using its native sampler and editing strategy, while applying our method under the same sampler but with our own attention control strategy. This setup ensures a fair comparison and demonstrates our method’s robustness across varying samplers. As shown in the table, the evaluation metrics of RF-Solver and FireFlow closely match those from fixed-seed generation, suggesting an inability to preserve structural consistency. Therefore, we exclude these two methods from subsequent comparisons on structure-consistent editing tasks. In contrast, our method consistently produces the best structure-preserving results.

\begin{table}[t]
\small
\centering
\caption{Quantitative results of \textit{Change Color} and \textit{Change Material} tasks.}
\resizebox{\linewidth}{!}{
\begin{tabular}{l|c|cc|cc}
\toprule
\multicolumn{1}{c|}{\multirow{2}{*}{Method}} & \multicolumn{1}{c|}{Canny}  & \multicolumn{2}{c|}{BG Preservation}  & \multicolumn{2}{c}{$\text{Clip Similarity}~\uparrow$}      \\
\cline{2-6}
\multicolumn{1}{c|}{} & $\text{SSIM}~\uparrow$ &  
PSNR~$\uparrow$ & 
SSIM~$\uparrow$ &
Whole & 
Edited
\\
\midrule
  SDEdit~\citeyearpar{meng2021sdedit} & 0.6795 & 23.99 & 0.8697 & 26.59 & 22.80 \\
  UniEdit-Flow~\citeyearpar{jiao2025uniedit} & 0.8029  & 30.56 & 0.9554 & 26.55 & 22.59 \\
  DiTCtrl~\citeyearpar{cai2024ditctrl} & 0.8235 & 29.54 & 0.9632 & 26.63 & 22.97 \\ 
  Ours & \textbf{0.8811} & \textbf{36.76} & \textbf{0.9869} & \textbf{27.19} & \textbf{23.73} \\
  \bottomrule 
\end{tabular}
}
\label{table:pie-bench}
\end{table}

Tab.~\ref{table:pie-bench} presents the whole benchmark with other baselines. Our method delivers superior results in both preserving source content and executing accurate edits, achieving \textbf{state-of-the-art} performance across the board.
\subsection{Qualitative Evaluation}
In this section, the evaluation begins with structure-consistent editing, highlighting the method’s ability to preserve structural consistency. This is followed by demonstrations on real images to validate practical effectiveness. Performance on structure-inconsistent editing is then presented, showcasing adaptability across varied scenarios. Finally, multi-round editing examples are provided, combining both structure-consistent and -inconsistent editing to demonstrate the method’s robustness and flexibility.

\subsubsection{Structure-consistent editing.} 
\begin{figure*}[t] 
  \centering
  \includegraphics[width=\textwidth]{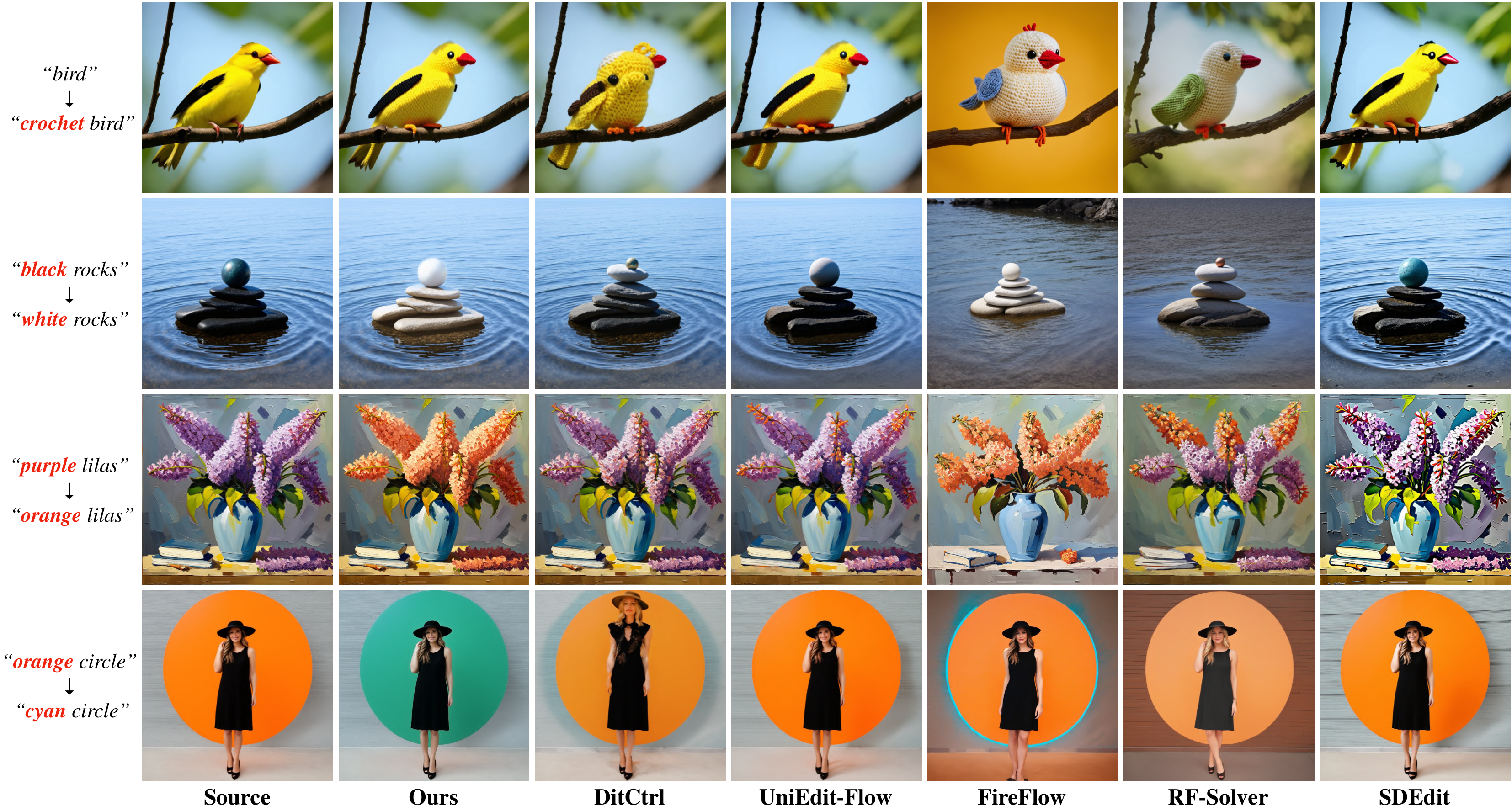}
  \caption{Qualitative comparison of methods on structure-consistent editing tasks.}
  \label{fig:consistent_compare}
\end{figure*}

Fig.~\ref{fig:consistent_compare} presents a qualitative comparison across all methods on structurally consistent editing tasks. Our approach accurately changes the color or material according to the target prompt while preserving the structure of the edited region same to that of the source image. Notably, beyond merely replacing colors, the edited outputs are also well adapted to the surrounding lighting conditions. In contrast, other methods often produce incorrect or insufficient edits and fail to maintain structural consistency. Additionally, our method faithfully preserves the non-edited regions, whereas others introduce undesirable changes. More results are shown in Appendix~\ref{sec:more_results}.

\subsubsection{Structure-consistent editing on real images.} We compare our real image editing results with several existing methods, including U-Net-based approaches such as FreePromptEditing~\cite{liu2024towards}, EditFriendly~\cite{huberman2024edit}, InfEdit~\cite{xu2023inversion}, PnP-Inv.~\cite{ju2023direct}, PnP~\cite{tumanyan2023plug}, and P2P~\cite{hertz2022prompt}. As shown in Fig.~\ref{fig:hat_compare}, conventional U-Net-based and MM-DiT-based methods all struggle to preserve the non-edited regions and often fail to accurately modify the hat color. In contrast, our method achieves the best performance, delivering precise edits in the target region while preserving the consistency of non-edited areas. Please refer to Appendix~\ref{sec:more_results} for further examples.

\subsubsection{Structure-inconsistent editing.} 
\begin{figure*}[!t] 
  \centering
  \includegraphics[width=1\textwidth]{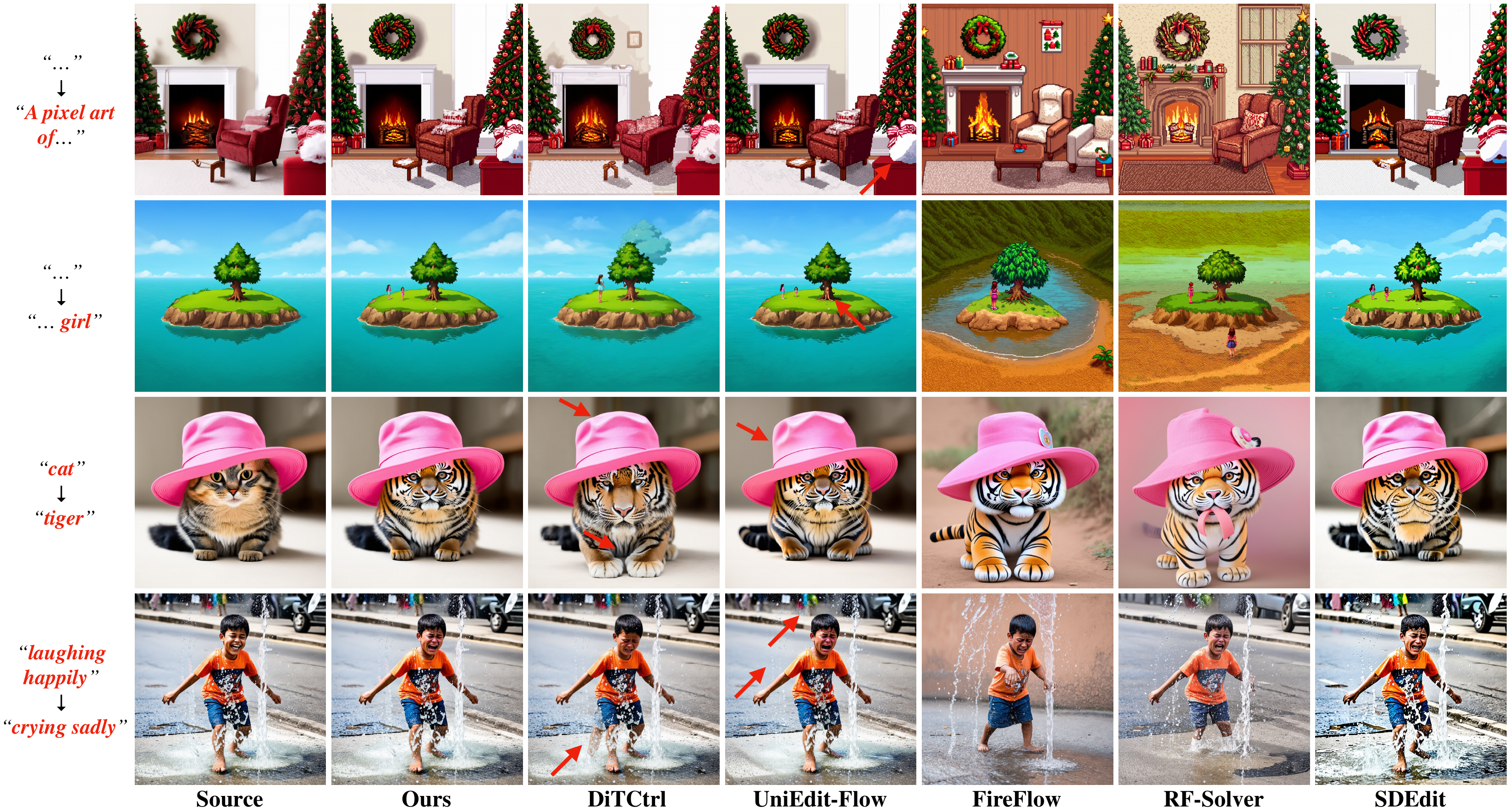}
  \caption{Qualitative comparison of methods on structure-inconsistent editing tasks.}
  \label{fig:inconsistent_compare}
\end{figure*}

We compare various methods on structure-inconsistent editing tasks in Fig.~\ref{fig:inconsistent_compare}. In these experiments, the consistency strength ($\alpha$) is set to 0.3, allowing the model to moderately edit structures for improved prompt alignment, while still preserving the overall layout. As shown, our method achieves better results in the edited regions, producing more precise editing with fewer artifacts. Moreover, it more effectively preserves the non-edited areas compared to other approaches, maintaining high content fidelity with respect to the source image. More results are shown in Appendix~\ref{sec:more_results}.

\subsubsection{Multi-round interactive editing on real images.} Fig.~\ref{fig:real_edit} presents an example of multi-round editing on a real input image. The image is first inverted into the latent space. Then, we perform a series of edits on it, including modifying the clothing color, motion, and hair. These flexible operations and the reliability of the results open up new possibilities for interactive or iterative editing tasks.

\subsection{Fine-grained Editing Control}

\begin{figure}[t]
  \centering
  \includegraphics[width=1\linewidth]{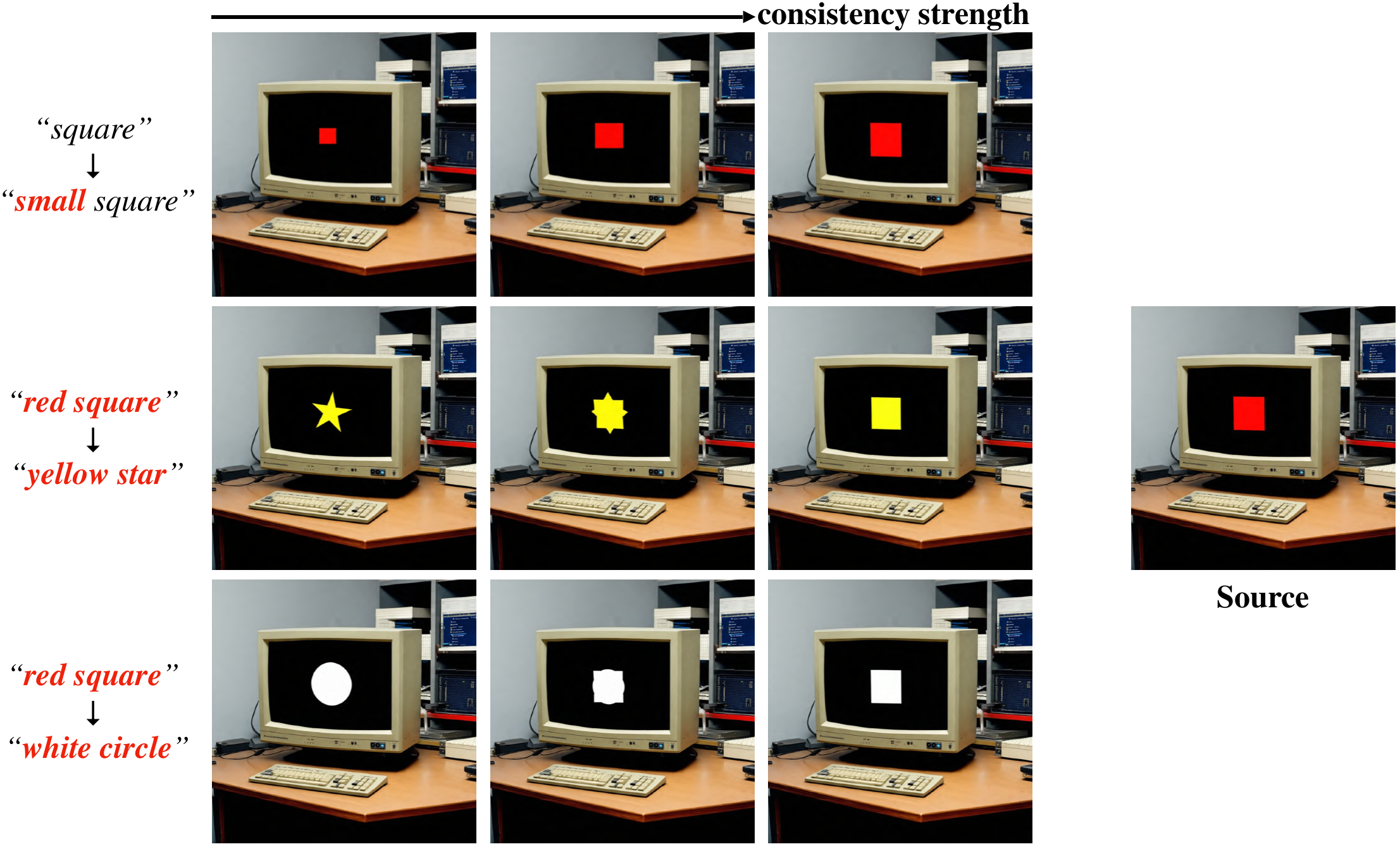}
  \caption{Effect of consistency strength on structural consistency. High strength strictly enforces structural preservation, while low strength permits prompt-driven shape changes. Texture editing remains consistent, highlighting effective disentanglement.}
  \label{fig:disentangle}
\end{figure}

Fig.~\ref{fig:disentangle} shows the controllability of structural consistency during editing by varying the consistency strength ($\alpha$). A high value enforces strict structure preservation, even when the prompt includes shape-altering instructions. At the same time, it still enables accurate texture editing, \textit{e.g.}, color changes, as specified by the prompt. In contrast, a low consistency strength permits structural editing with the prompt. Additionally, the similar color appearance under varying consistency strengths demonstrates the effectiveness of our disentangled structure-preserving control mechanism, enabling precise and independent editing of structure and texture.


\begin{figure}[t]
  \centering
  \includegraphics[width=0.93\linewidth]{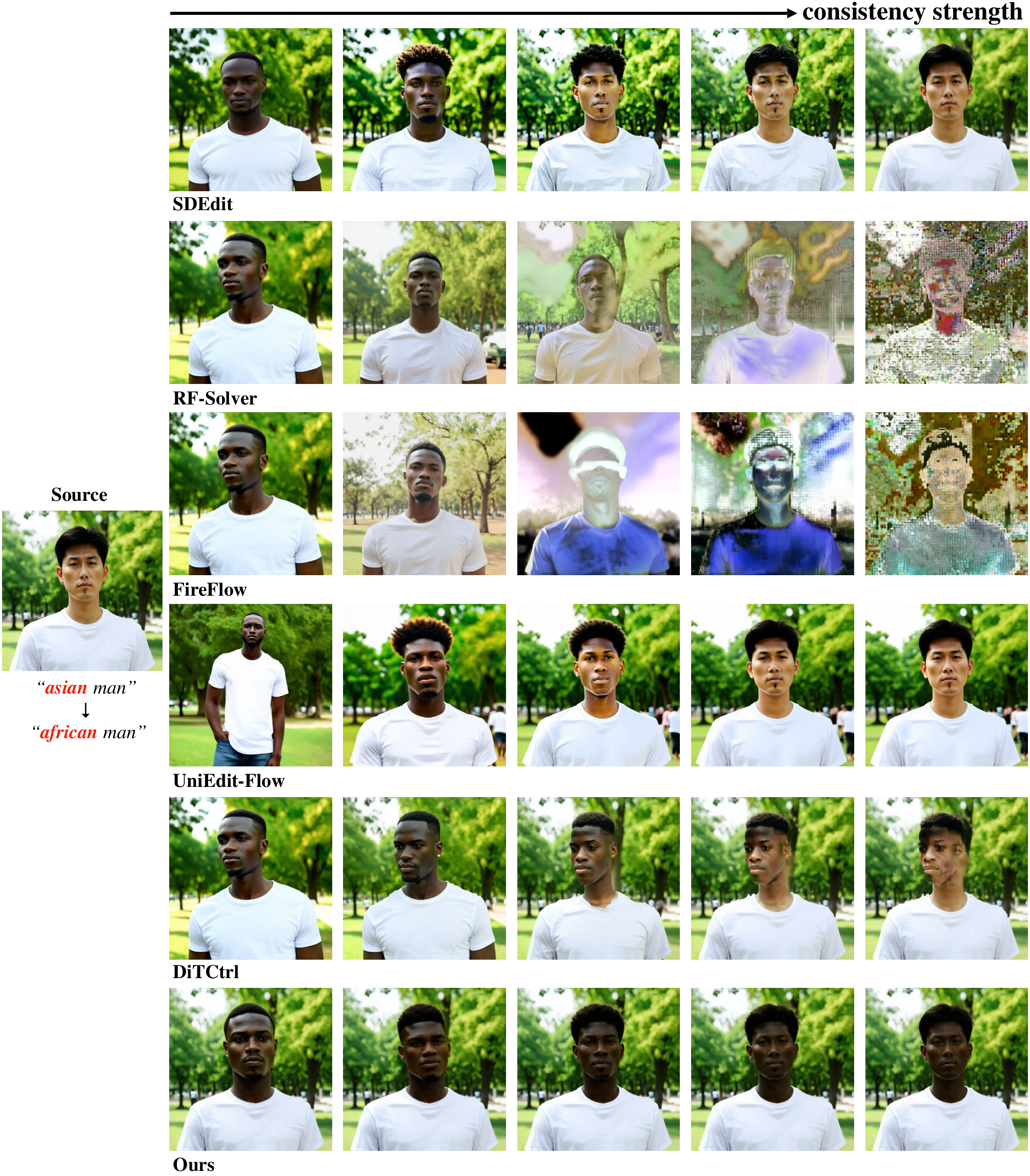}
  \caption{ Qualitative comparison on consistency strength adjustment. }
  \label{fig:edit_strength_compare}
\end{figure}

Furthermore, thanks to this disentanglement, our method enables smooth and controllable adjustment of consistency strength. In contrast, other methods struggle to maintain stable editing performance across varying strength levels, often relying on specific parameter values, as illustrated in Fig.~\ref{fig:edit_strength_compare}. This property further highlights the potential for integrating a controllable consistency strength slider into interactive editing interfaces.

\subsection{Ablation}

We conduct two ablation studies to validate the effectiveness of our approach.

\begin{figure}[t]
  \centering
  \includegraphics[width=0.95\linewidth]{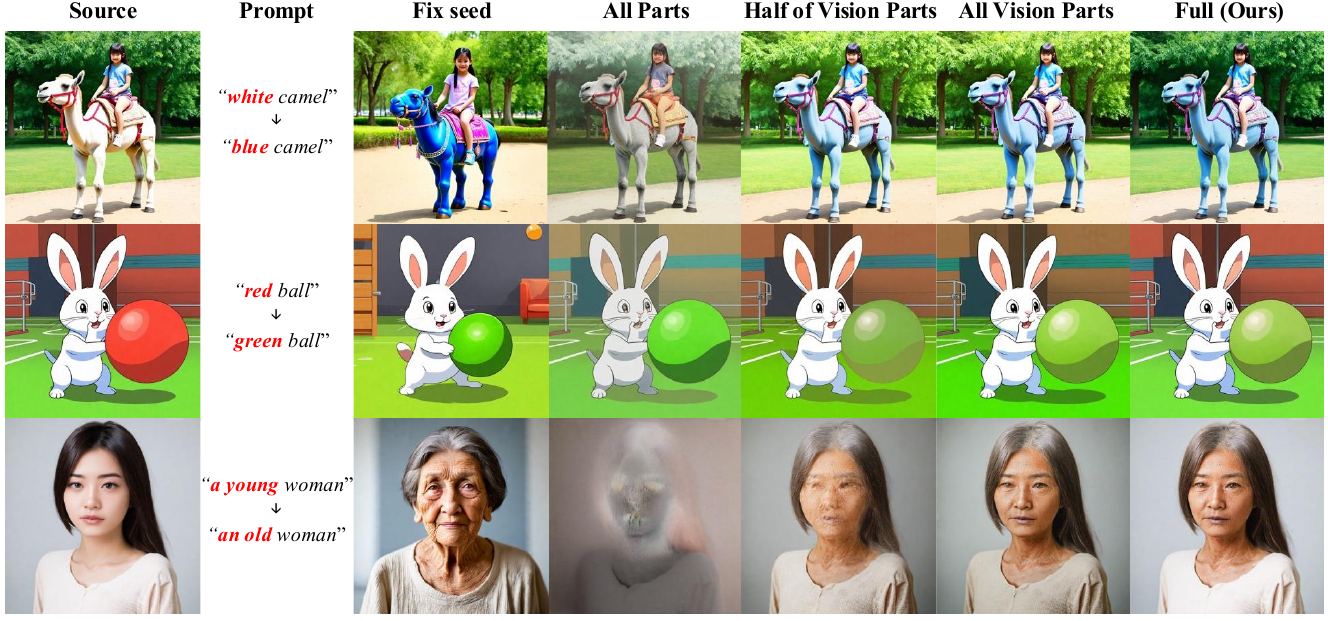}
  \caption{Ablation study on attention control for structure consistency. We compare (1) fixed seed results, (2) swapping all \textit{Q} and \textit{K} tokens across all blocks, (3) swapping only the vision part of \textit{Q} and \textit{K} tokens in the last half of the blocks, (4) swapping only the vision part of \textit{Q} and \textit{K} tokens in all blocks, and (5) adding our non-editing region consistency module.}
  \label{fig:ablation_swap_layer}
\end{figure}

\subsubsection{Structural consistency in edited regions.} As shown in Fig.~\ref{fig:ablation_swap_layer}, we conduct an ablation study to investigate the effects of different $\boldsymbol{Q}$ and $\boldsymbol{K}$ tokens swapping strategies. Starting from the same seed ensures a well-initialized structural layout for subsequent editing. Swapping all (text and vision) $\boldsymbol{Q}$ and $\boldsymbol{K}$ tokens preserves structural consistency to a certain extent but significantly impairs text-driven editability, as it discarding the text tokens of target. In contrast, selectively swapping only the vision part of $\boldsymbol{Q}$ and $\boldsymbol{K}$ tokens across all blocks maintains the structural layout of the source image while preserving strong editing capabilities. To verify the necessary of swapping in all layers, we find that only swap the latter half of the model’s blocks will substantially weaken structural control and can lead to corrupted generation results. Finally, by incorporating our content fusion method on top of the full-block vision-only $\boldsymbol{Q}$ and $\boldsymbol{K}$ swapping, we further enforce preservation in non-edited regions, achieving the best quality. These findings emphasize the importance of applying editing across all blocks while restricting editing to the vision parts of the attention mechanism.

\begin{table}[t]
\small
\centering
\caption{ Evaluation of content preservation in non-edited regions.}
\begin{tabular}{l|cccc}
\toprule
\multicolumn{1}{c|}{} & \textcolor{lightgray}{DiffEdit~\citeyearpar{couairon2022diffedit}} & $\boldsymbol{V}$ & $\boldsymbol{Q}$ \& $\boldsymbol{K}$ & $\boldsymbol{Q}$ \& $\boldsymbol{K}$ \& $\boldsymbol{V}$~(Ours) \\
\midrule
  $\text{PSNR}\uparrow$ & \textcolor{lightgray}{51.49} & 37.98  &  24.32 & \textbf{38.85}    \\
  $\text{SSIM}~\uparrow$ & \textcolor{lightgray}{0.9972} &  0.9905  &  0.9286 & \textbf{0.9917}    \\
  \bottomrule 
\end{tabular}
\label{table:background-compare}
\end{table}

\begin{figure}[t]
  \centering
  \includegraphics[width=\linewidth]{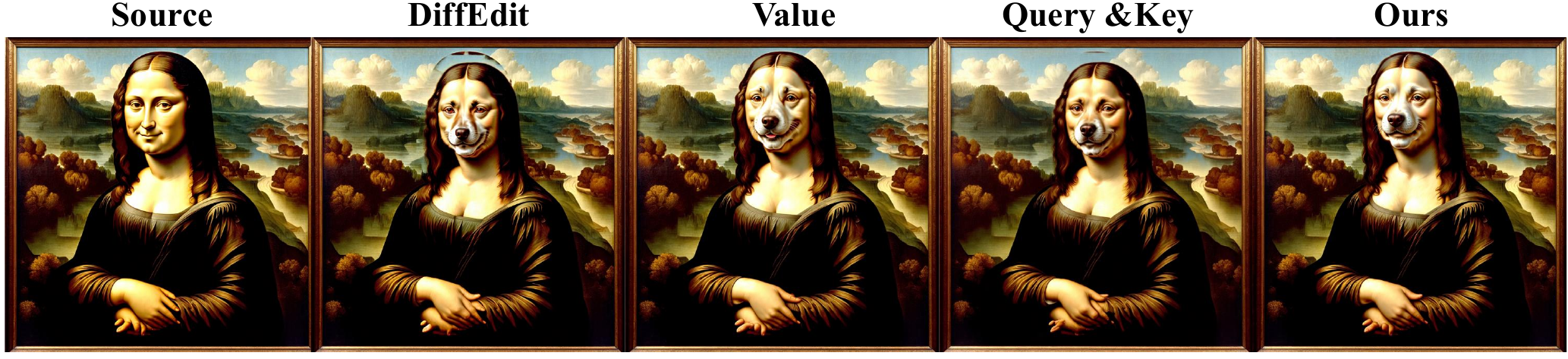}
  \caption{Ablation study of non-edited region preservation. The edit prompt is ``a head'' $\to$ ``a dog head''.}
  \label{fig:ablation_bg}
\end{figure}

\subsubsection{Content preservation in non-edited regions.}
Tab.~\ref{table:background-compare} reports PSNR and SSIM scores on the non-edited regions of 80 image pairs from the \textit{Change Color} and \textit{Change Material} tasks in PIE-Bench~\cite{ju2023direct}, evaluating how well different methods preserve content consistency in non-edited regions. All methods use the same binary mask, extracted using our mask extraction method. According to the results in Fig.~\ref{fig:ablation_bg}, we can see a hard replacement strategy described in DiffEdit~\cite{couairon2022diffedit} introducing visible artifacts at transition boundaries. Secondly, swapping only the vision tokens of $\boldsymbol{Q}$ and $\boldsymbol{K}$ maintains structural consistency but introduces slight color shifts, which degrade metric scores in Tab.~\ref{table:background-compare}. In contrast, swapping only the vision part of the $\boldsymbol{V}$ tokens yields a more stable preservation. Finally, Tab.~\ref{table:background-compare} and Fig.~\ref{table:background-compare} shows that combining vision-token swaps of $\boldsymbol{Q}$, $\boldsymbol{K}$, and $\boldsymbol{V}$ achieves the best results in both quantitatively and qualitatively, as it preserves more details.

\subsection{Compatibility and Application}

\begin{figure}[t] 
  \centering
  \includegraphics[width=\linewidth]{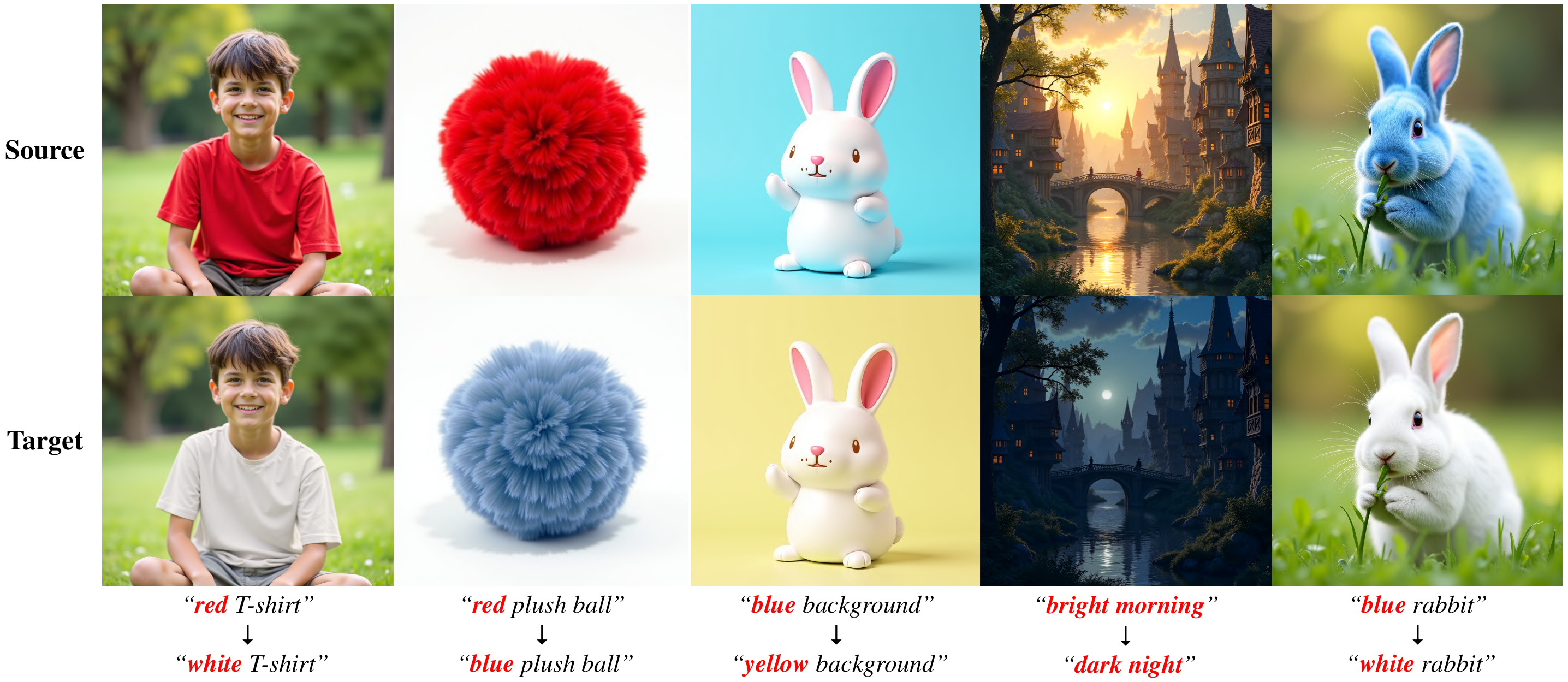}
  \caption{Examples of editing results with FLUX.}
  \label{fig:flux}
\end{figure}

\subsubsection{Generalization to MM-DiT variants.} 
Our method not only works effectively with SD3 but also generalizes well to other MM-DiT variants such as FLUX.1-dev~\cite{flux2024}. In Fig.~\ref{fig:flux}, the consistent preservation of fine-grained details and the accurate adaptation of lighting-related reflections further highlight the potential of our approach when applied to more powerful future models.

\begin{figure}[t]
  \centering
  \includegraphics[width=\linewidth]{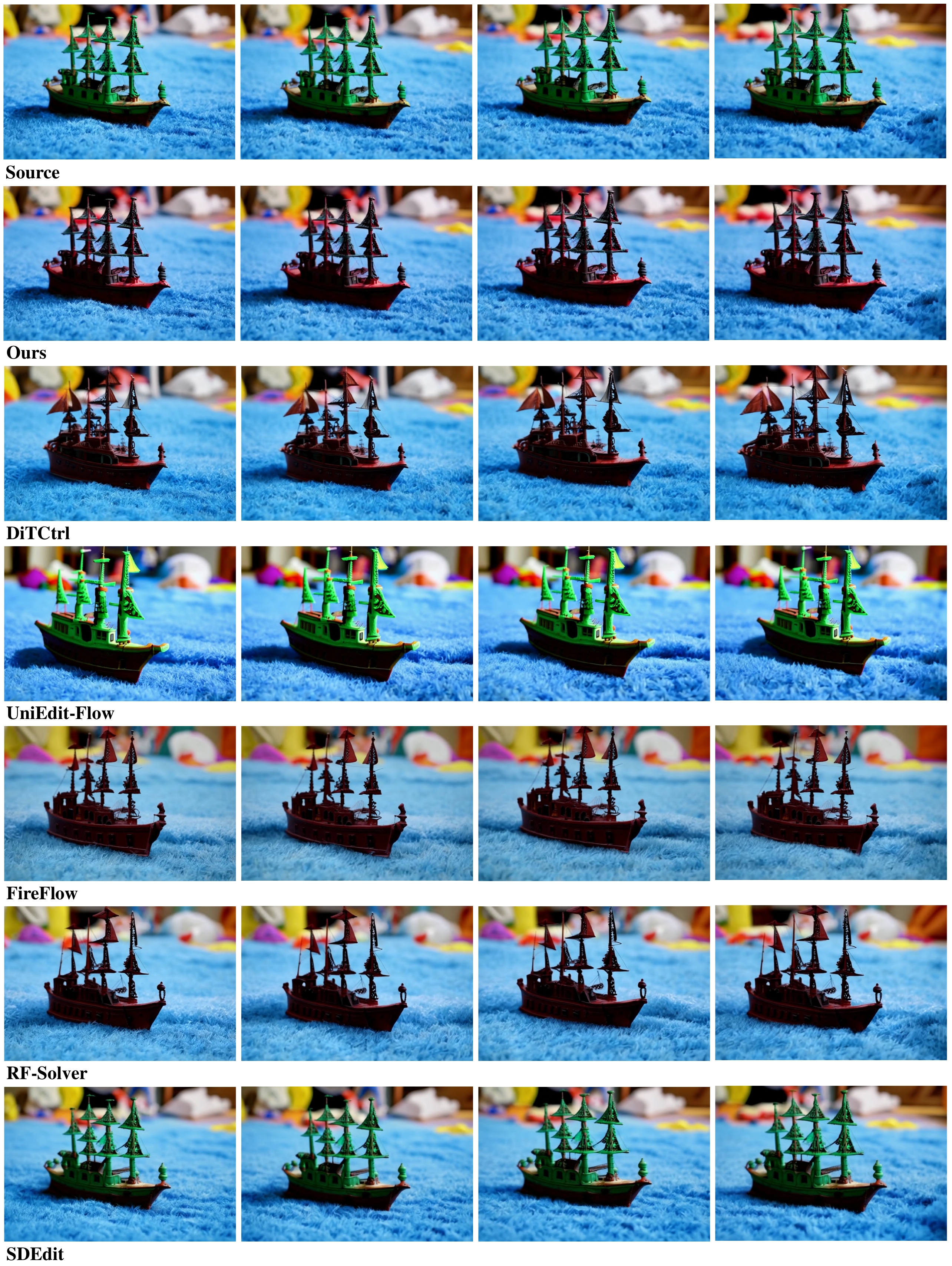}
  \caption{Qualitative comparison of methods on video editing tasks. The edit prompt is ``green toy ship'' $\to$ ``dark red toy ship''. }
  \label{fig:video_compare}
\end{figure}

\subsubsection{Generalization to video editing.}
While our method has already been demonstrated to be agnostic to specific samplers, we further showcase its broad applicability across generation methods (\textit{e.g.}, diffusion models) and domains (\textit{e.g.}, video) by applying it to CogVideoX-2B~\cite{yang2024cogvideox}, a diffusion-based video generation model. As shown in Fig.~\ref{fig:video_compare}, our approach enables consistent and controllable editing in both the spatial and temporal domains. Importantly, small inconsistencies that may go unnoticed in static images often become amplified and distracting in videos. Our method effectively highlights its robustness and generalizability. Additional results are provided in Appendix~\ref{sec:more_results}.

\begin{figure}[t] 
  \centering
  \includegraphics[width=\linewidth]{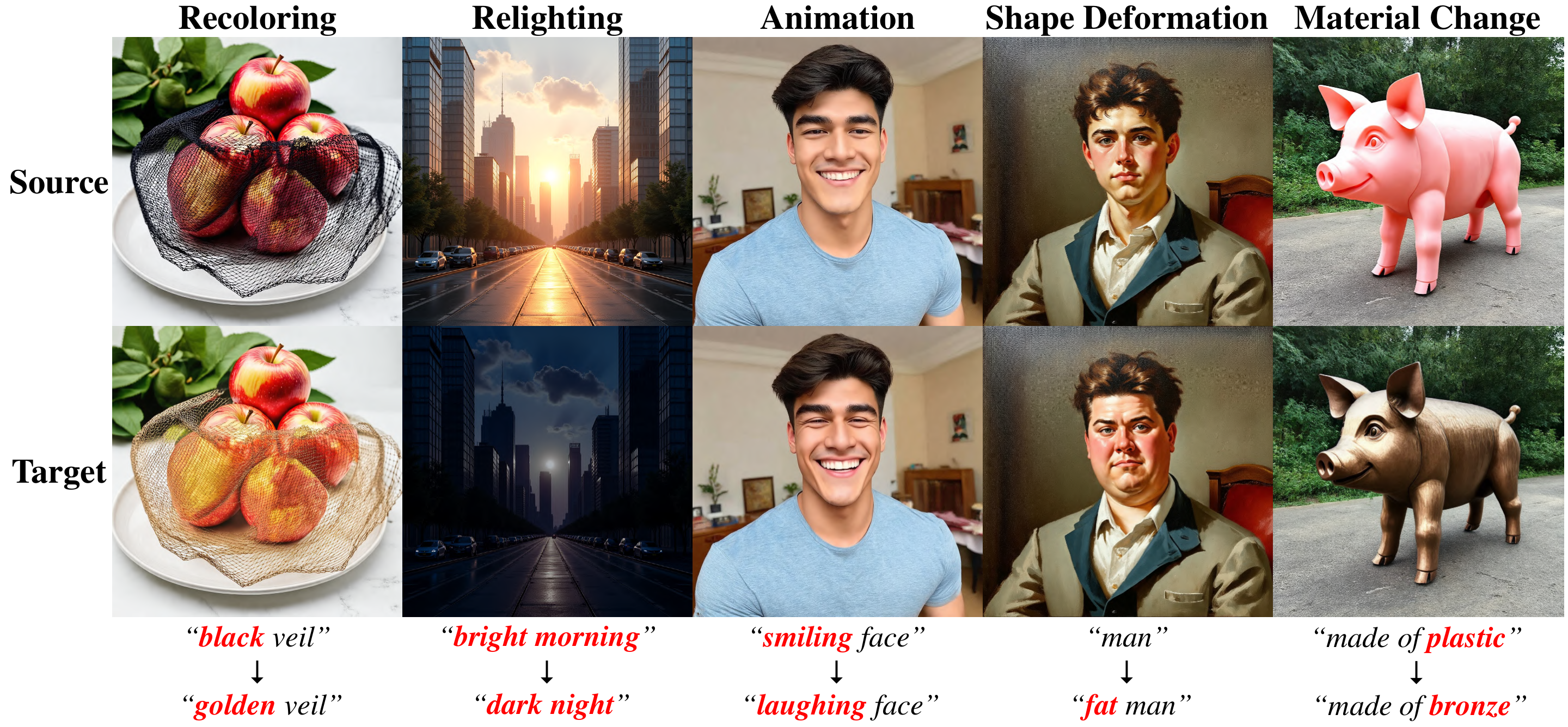}
  \caption{Examples of applications.}
  \label{fig:application}
\end{figure}

\subsubsection{Application.}
Fig.~\ref{fig:application} showcases our method’s versatility across several challenging editing tasks, including recoloring, relighting, animation, shape deformation, and material change. Extending these capabilities to video further amplifies creative possibilities by enabling temporally consistent and detailed edits. The strong editing power and ease of use highlight the broad potential of our approach for practical and scalable content creation.

\section{Conclusion}

In this work, we identify key limitations of existing training-free editing methods, including their inability to achieve both strong and consistent text-guided editing, as well as their lack of fine-grained control, where most prior approaches were designed for U-Net or naively applied to MM-DiT without architectural adaptation. To address this, we conduct a detailed analysis of the attention mechanism in MM-DiT and uncover three critical insights that reveal why existing methods fall short. Building on these findings, we propose ConsistEdit, a novel attention control method that operates exclusively on vision tokens. By separating editing and non-editing regions and applying differentiated attention manipulation, ConsistEdit achieves precise, structural consistent edits in edited regions while preserving content in non-edited regions.

Extensive experiments demonstrate that ConsistEdit achieves state-of-the-art performance across diverse image and video editing tasks, without requiring manual tuning. It delivers reliable performance out of the box while offering users fine-grained control over structural consistency. These findings highlight the potential of MM-DiT when paired with our attention control strategies.

\begin{acks}
We are deeply grateful to Morph Studio and its CEO, Huaizhe Xu, for their generous provision of the computational resources that made this work possible. We would also like to thank Heung-Yeung Shum, Baoyuan Wang, Lei Zhang, Xuan Ju, Guanlong Jiao, Yukai Shi, Shihao Zhao, and Bojia Zi for their valuable discussions and insightful feedback throughout the course of this research.
\end{acks}

\bibliographystyle{ACM-Reference-Format}
\bibliography{reference}

\clearpage

\appendix

\section{Implementation Details}
\label{sec:imp_details}

\subsection{Inference Settings}

We use 28 inference steps for SD3~\cite{esser2024scaling}, 50 for CogVideoX-2B~\cite{yang2024cogvideox}, and 20 for FLUX.1-dev~\cite{flux2024}. The classifier-free guidance (CFG) scale~\cite{ho2022classifier} is set to 7.5 for editing generated sources and 2.0 for real-image editing. All images are generated at $1024 \times 1024$, and all videos at $720 \times 480$. The inference device is $1\times$ RTX-4090 GPU.

Our method is compatible with various samplers, with the Euler sampler adopted as default unless otherwise specified. It also supports various inversion techniques; in all experiments, we use the latest inversion method from UniEdit-Flow~\cite{jiao2025uniedit}. We fix the consistency strength to $\alpha = 1$ for tasks requiring structural preservation, and set it to $\alpha=0.3$ for other tasks.

We adopt a default mask threshold of $0.1$, which consistently performs well across our experiments. This relatively coarse masking suffices thanks to the generation models' strong global adaptation, allowing it to propagate edits from partial color cues to semantically aligned regions. The target object for mask extraction is identified either using ``blended\_word'' keywords from PIE-Bench~\cite{ju2023direct}, or simply by extracting the noun of the object to be edited. Furthermore, our method supports externally provided masks, enabling users to integrate masks generated from other pipelines for more flexible control.

\subsection{Sampling Details}

To accelerate inference and reduce the number of function evaluations (NFE) during sampling, similar to the approach in~\citet{wang2024taming}, we first run the source prompt branch and cache the \textbf{\textit{Q}}, \textbf{\textit{K}}, and \textbf{\textit{V}} tokens at each step and block for later use. During this stage, we also compute and store the final averaged editing mask. When editing with the target prompt, we load the stored \textbf{\textit{Q}}, \textbf{\textit{K}}, and \textbf{\textit{V}} tokens from the source and apply the editing mask through Mask-Guided Attention Fusion as needed. This strategy ensures that the mask extraction and editing process introduces no additional NFE, maintaining the same efficiency as standard sampling methods.

\subsection{Implementation of Compared Methods}

Since some compared methods do not provide implementations for SD3 or CogVideoX-2B, or compatible sampling code, we re-implement them within the SD3 and CogVideoX-2B framework by faithfully following their original logic and carefully tuning hyperparameters to match the reported performance. Implementation details are as follows:

\begin{itemize}[leftmargin=*, itemsep=0pt]
\item \textbf{DiTCtrl~\cite{cai2024ditctrl}}: For image editing, we set the start timestep to $2$ and the end timestep to $17$, applying edits to the last $5$ blocks. For video editing, we use the official implementation. During the editing steps, \textbf{\textit{K}} and \textbf{\textit{V}} tokens are copied from the source branch to the target branch in the attention layers. For this method, consistency strength is controlled by increasing the number of end step during which \textbf{\textit{K}} and \textbf{\textit{V}} tokens are shared.

\item \textbf{UniEdit-Flow~\cite{jiao2025uniedit}}: The official implementation is based on SD3, but only provides the parameter $\omega$ for CFG = $1$. Following the similarity transformation described in the paper, we adopt $\omega = 5 \div 7.5 \approx 0.6$ and set $\alpha = 0.6$, which yields performance comparable to the original. The same settings are used for video generation. Under this setting, consistency strength is controlled by decreasing the value of $\alpha$.

\item \textbf{FireFlow~\cite{deng2024fireflow}}: We observe a significant drop in source–target consistency as CFG increases. Due to performance degradation when the number of edited timesteps increases (as shown in Fig.~\ref{fig:attn_fusion}), we limit editing to timesteps from $0$ to $3$ across all blocks. For video generation, the end timestep is set to $9$. During editing, \textbf{\textit{V}} tokens are copied from source to target. In this approach, consistency strength is modulated by increasing the number of final step in which the \textbf{\textit{V}} features are shared.

\item \textbf{RF-Solver~\cite{wang2024taming}}: Similar to FireFlow, we set the editing range from timestep $0$ to $7$ for the latter half of the blocks. During editing, \textbf{\textit{V}} tokens are copied from the source to the target branch. For video generation, the end timestep is set to $9$. The end step of sharing of \textbf{\textit{V}} tokens serves as the control mechanism for consistency strength in this method.

\item \textbf{SDEdit~\cite{meng2021sdedit}}: We set $t_0 = 0.6$ and apply editing to either generated source content or real input content, for both image and video generation tasks. For this method, consistency strength is controlled by decreasing the value of $t_0$.
\end{itemize}

\subsection{Implementation of FLUX}
FLUX~\cite{flux2024} is composed of several double blocks and single blocks. As noted by~\citet{wang2024taming}, single blocks primarily encode general information relevant to generation. Therefore, we apply our editing methods specifically to the single blocks.

\section{Results and Analysis}
\subsection{More Results}
\label{sec:more_results}

\begin{figure*}[t!]
  \centering
  \includegraphics[width=\textwidth]{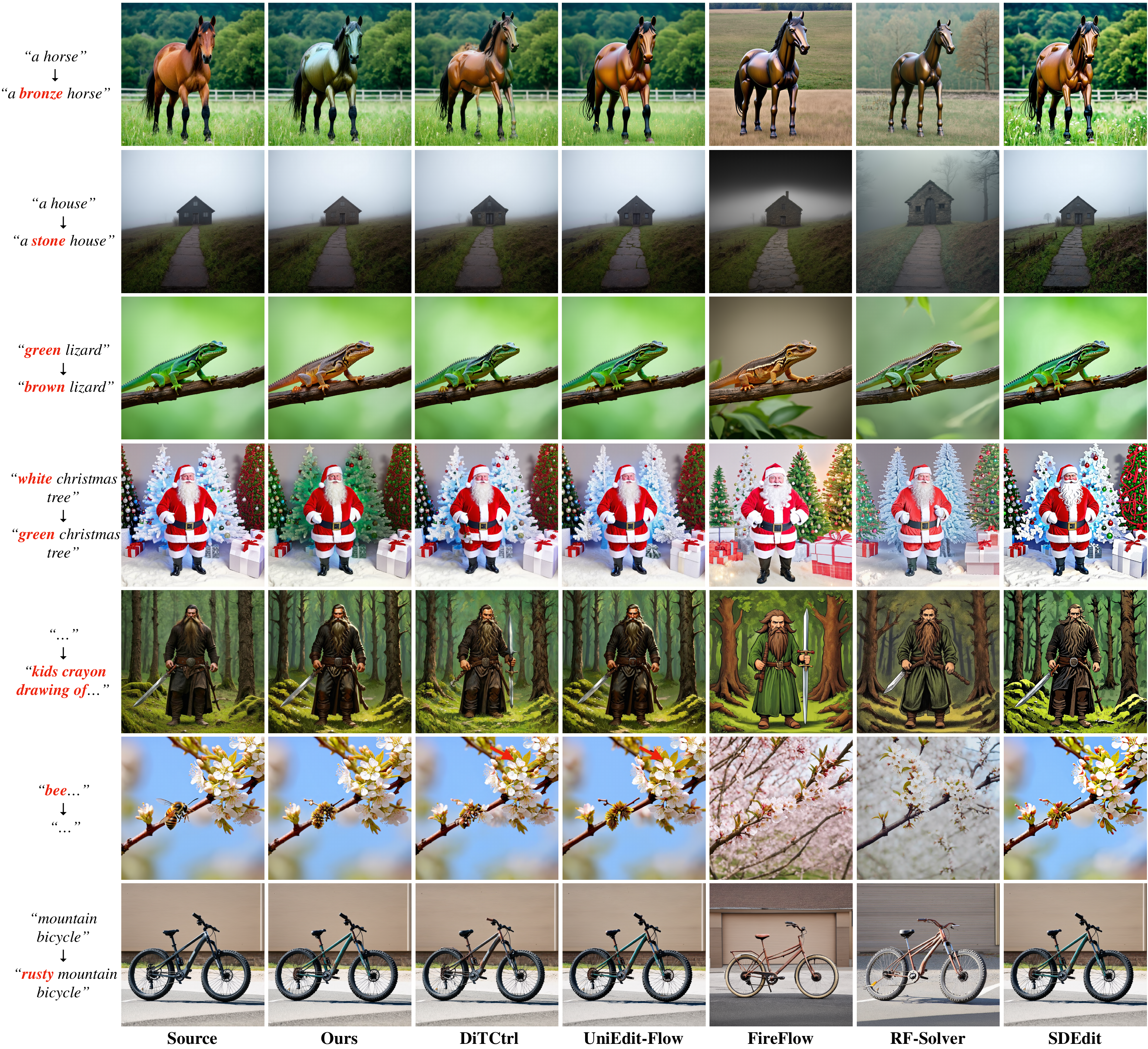}
  \caption{ Additional qualitative comparison of methods on structure-consistent and structure-inconsistent editing tasks. }
  \label{fig:more_image_compare}
\end{figure*}

Additional image editing comparisons are presented in Fig.~\ref{fig:more_image_compare}, covering both structure-consistent and structure-inconsistent editing tasks. The results demonstrate that our method achieves superior structural consistency, better preservation of non-edited regions, and enhanced editability compared to existing approaches.

\begin{figure}[t]
  \centering
  \includegraphics[width=\linewidth]{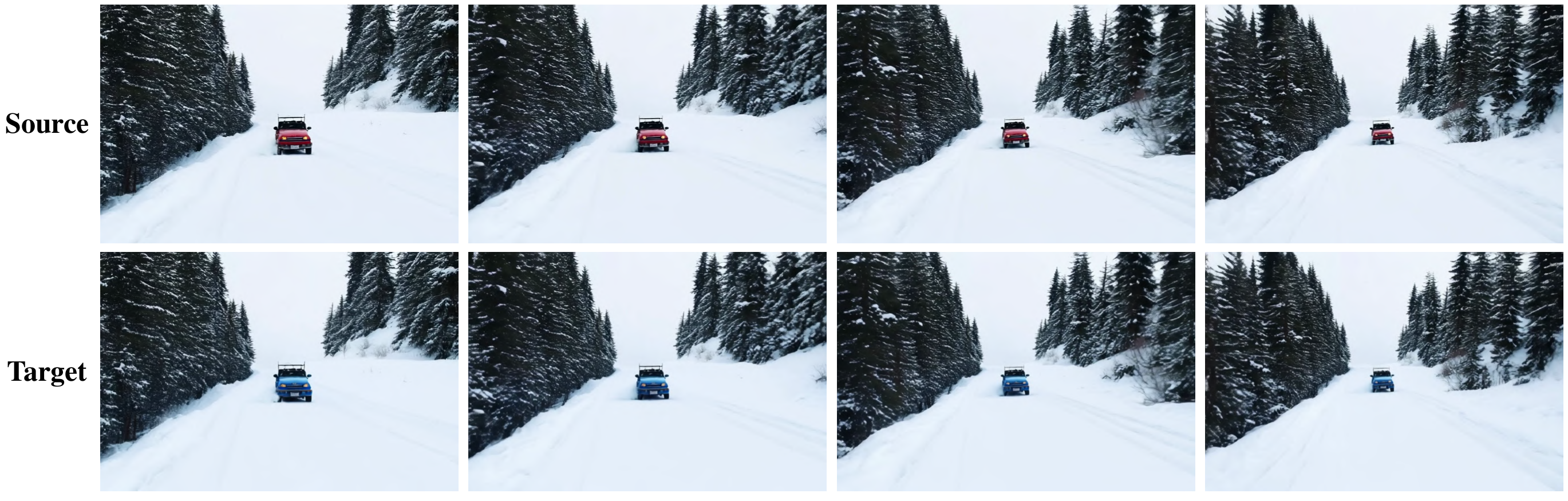}
  \caption{ Examples of video editing.. The prompt is ``red SUV'' $\to$ ``blue SUV''. }
  \label{fig:more_video}
\end{figure} %

\begin{figure}[h]
  \centering
  \includegraphics[width=\linewidth]{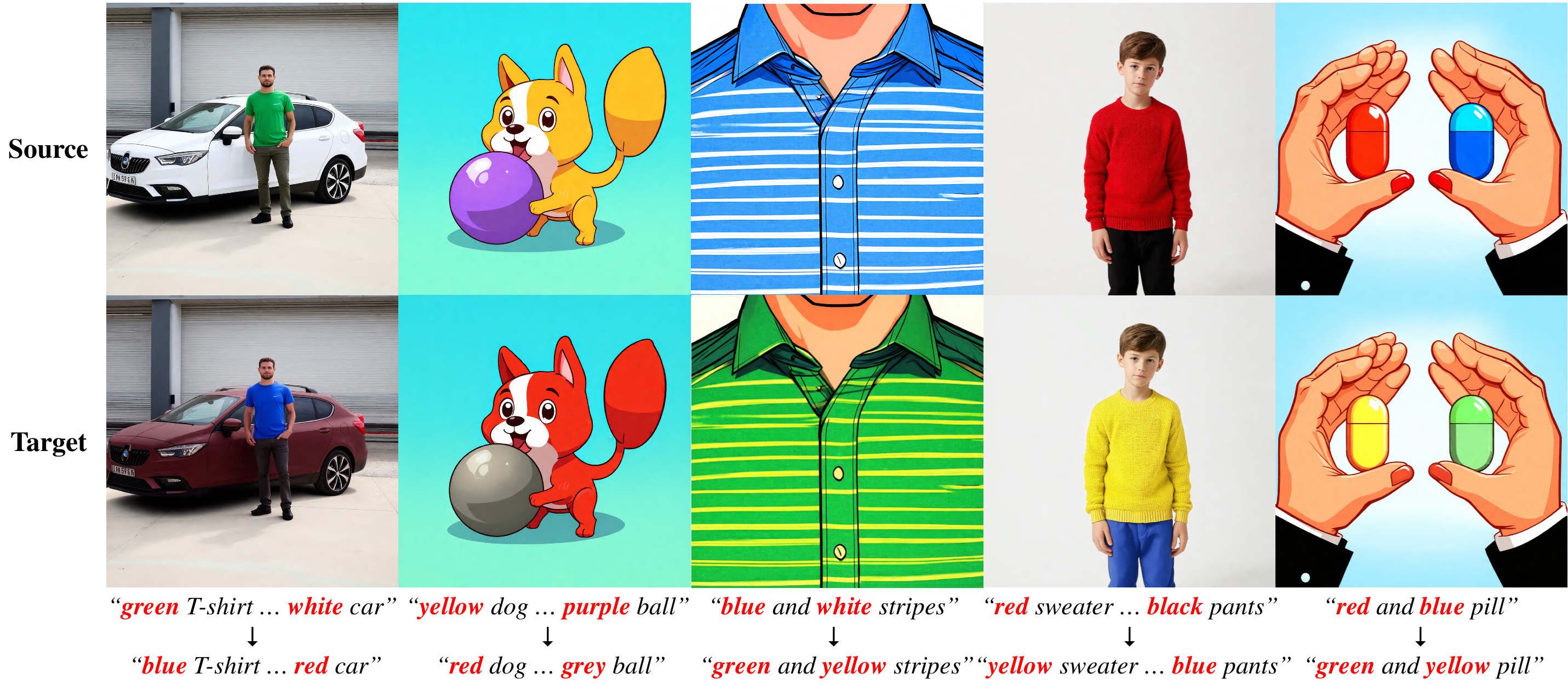}
  \caption{ Examples of multi-region editing.}
  \label{fig:multi_region}
\end{figure} %
\begin{figure}[h]
  \centering
  \includegraphics[width=\linewidth]{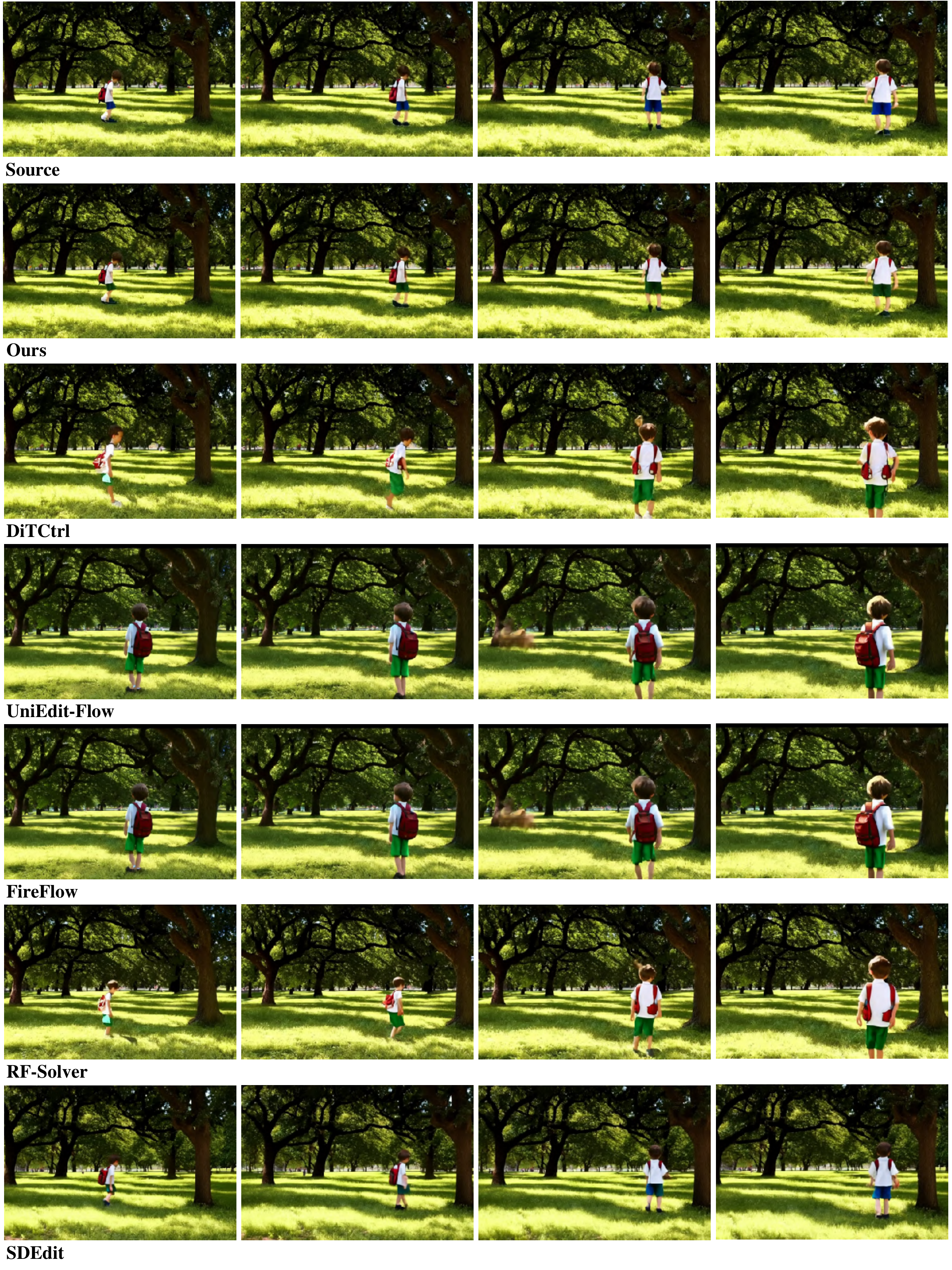}
  \caption{ Additional qualitative comparison of methods on video editing tasks. The edit prompt is ``blue shorts'' $\to$ ``green shorts''. }
  \label{fig:video_compare_2}
\end{figure} %
We present additional results on video editing tasks in Fig.~\ref{fig:more_video} and~\ref{fig:video_compare_2}. Fig.~\ref{fig:more_video} showcases examples generated by our method alone, while Fig.~\ref{fig:video_compare_2} provides comparisons with existing approaches, demonstrating our superior performance, particularly in scenarios with complex motion.

Fig.~\ref{fig:multi_region} presents additional cases of multi-region editing, demonstrating that our method can handle multi-object editing even in the presence of occlusion or complex geometric relationships. Notably, even when multiple regions exhibit intertwined textures, our method accurately identifies the target color for each region and performs the corresponding edits. These results highlight the precise text-driven control of our method, fine-grained understanding of visual structure, and strong structure preservation capabilities.

\begin{figure}[h!]
  \centering
  \includegraphics[width=\linewidth]{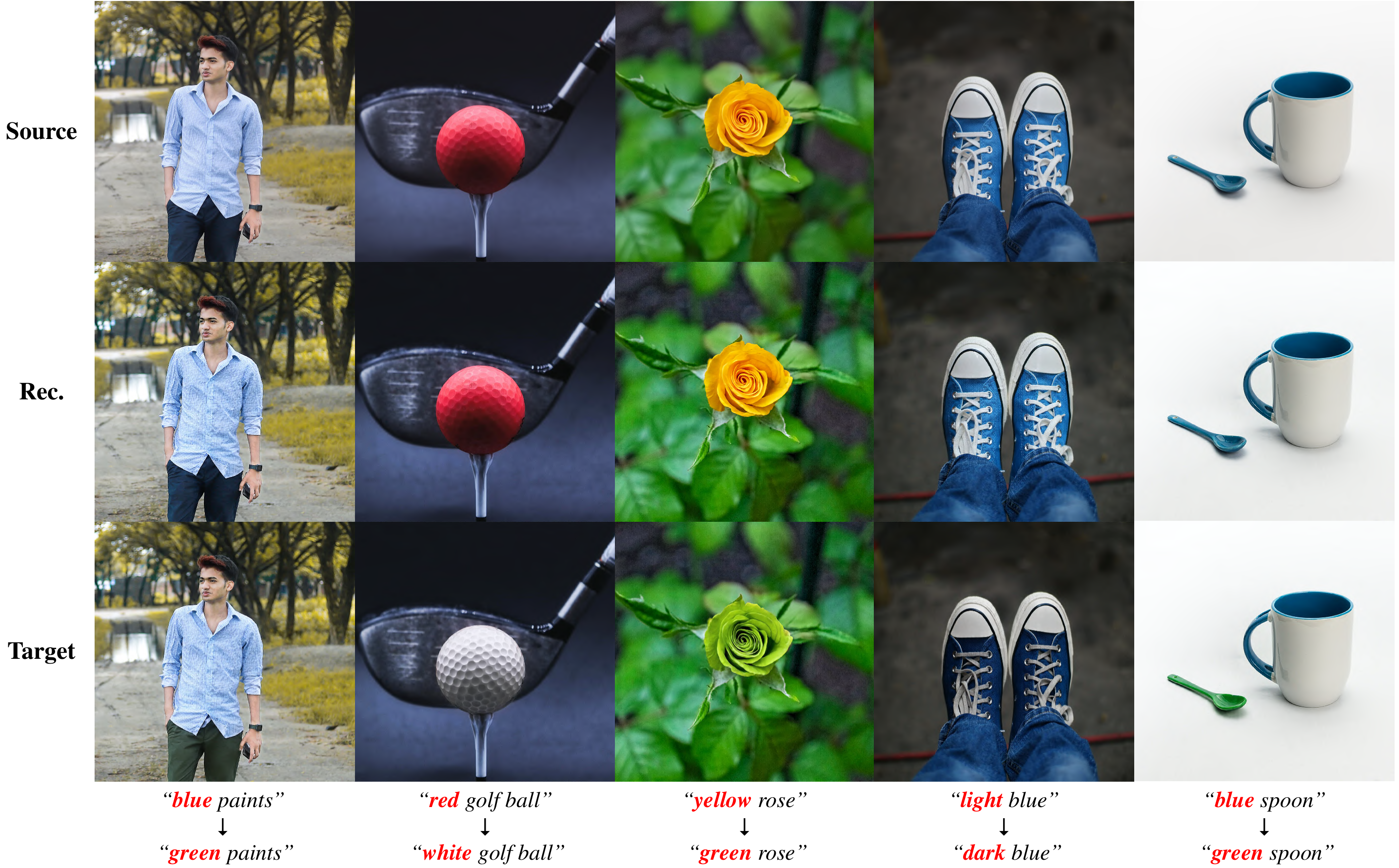}
  \caption{ Examples of real input image editing. The first row shows the source images, the second row presents the reconstructed images via inversion, and the third row displays the editing results based on the target prompts.}
  \label{fig:more_real_edit}
\end{figure} %

Additional real-image editing examples are shown in Fig.~\ref{fig:more_real_edit}. Our method preserves the structural integrity within the edited regions while maintaining the original content in the non-edited regions, achieving performance on par with editing generated images.

\subsection{User Study}
We conducted a user study involving 18 participants to evaluate editing quality across different methods. 
Each participant was presented with 30 randomly selected pairs of structure-consistent and structure-inconsistent edits, and was asked to choose the preferred result in each pair. 
As summarized in Tab.~\ref{tab:user_study}, Ours achieved a dominant preference rate of \textbf{71.11\%}, substantially outperforming all competing approaches. 

\begin{table}[t]
\centering
\begin{tabular}{l|c}
\toprule
\textbf{Method} & \textbf{Preference (\%)} \\
\midrule
RF-Solver~\cite{wang2024taming}      & 0.74  \\
SDEdit~\cite{meng2021sdedit}         & 5.19  \\
FireFlow~\cite{deng2024fireflow}       & 5.93  \\
UniEdit-Flow~\cite{jiao2025uniedit}   & 6.67  \\
DiTCtrl~\cite{cai2024ditctrl}        & 10.37 \\
Ours    & \textbf{71.11} \\
\bottomrule
\end{tabular}
\caption{User study preferences over different methods.}
\label{tab:user_study}
\end{table}

\subsection{Consistency Strength}

\begin{figure}[h!]
  \centering
  \includegraphics[width=\linewidth]{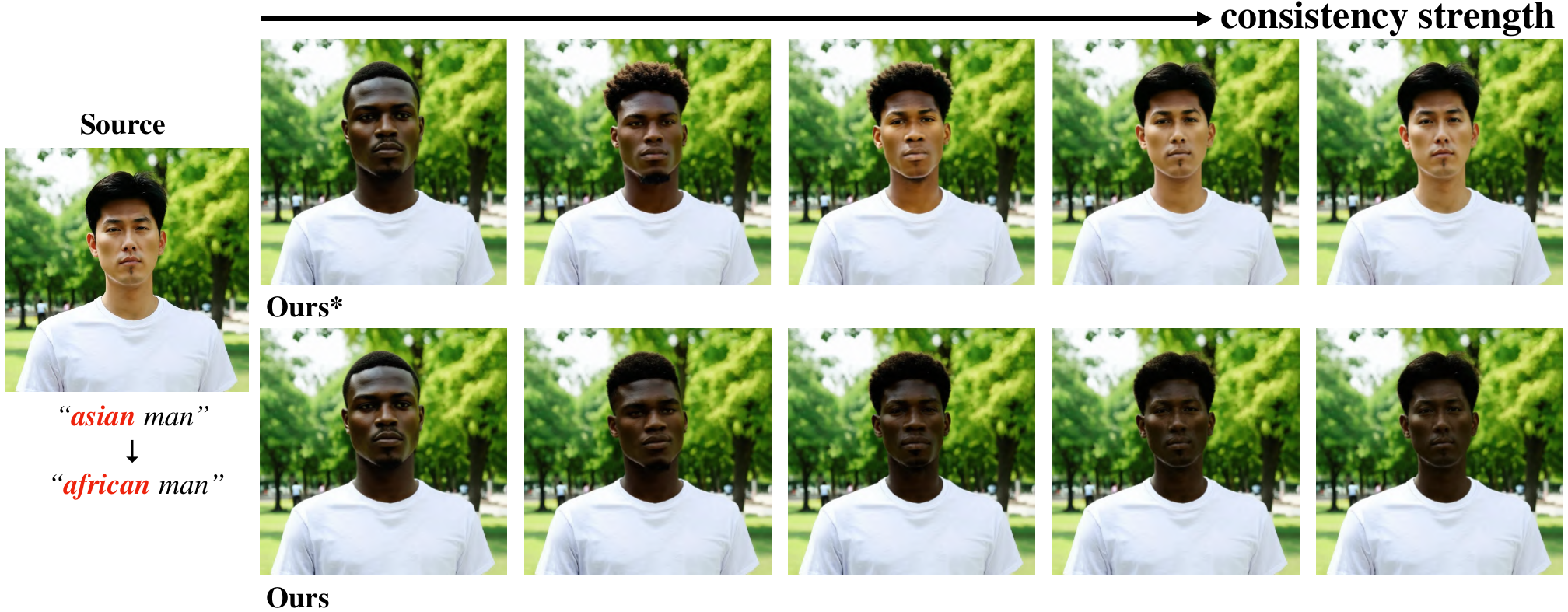}
  \caption{Qualitative comparison of different consistency strength settings. ``Ours'' denotes the method proposed in the main paper, while ``Ours*'' refers to a modified version of our method. }
  \label{fig:same_consist}
\end{figure} %

The main text demonstrates that our method offers fine-grained control over structural alignment with the source image through the consistency strength, while preserving the ability to edit texture according to the prompt. However, in certain downstream applications, users may prefer a binary behavior in which a consistency strength of $1$ results in an output identical to the source image, and a strength of $0$ produces results fully aligned with the edited prompt. Although such scenarios are beyond the primary focus of this work, we provide a simple mechanism to enable this behavior, which may serve as a basis for future research in this direction.

To support this behavior, we apply a small modification to our method: within the editing region, in addition to transferring the vision part of \textbf{\textit{Q}} and \textbf{\textit{K}} tokens, we also transfer that of \textbf{\textit{V}} tokens. As shown in Fig.~\ref{fig:same_consist}, this simple adjustment successfully achieves the desired behavior between unedited and fully edited results.

\begin{figure}[h]
  \centering
  \includegraphics[width=0.8\linewidth]{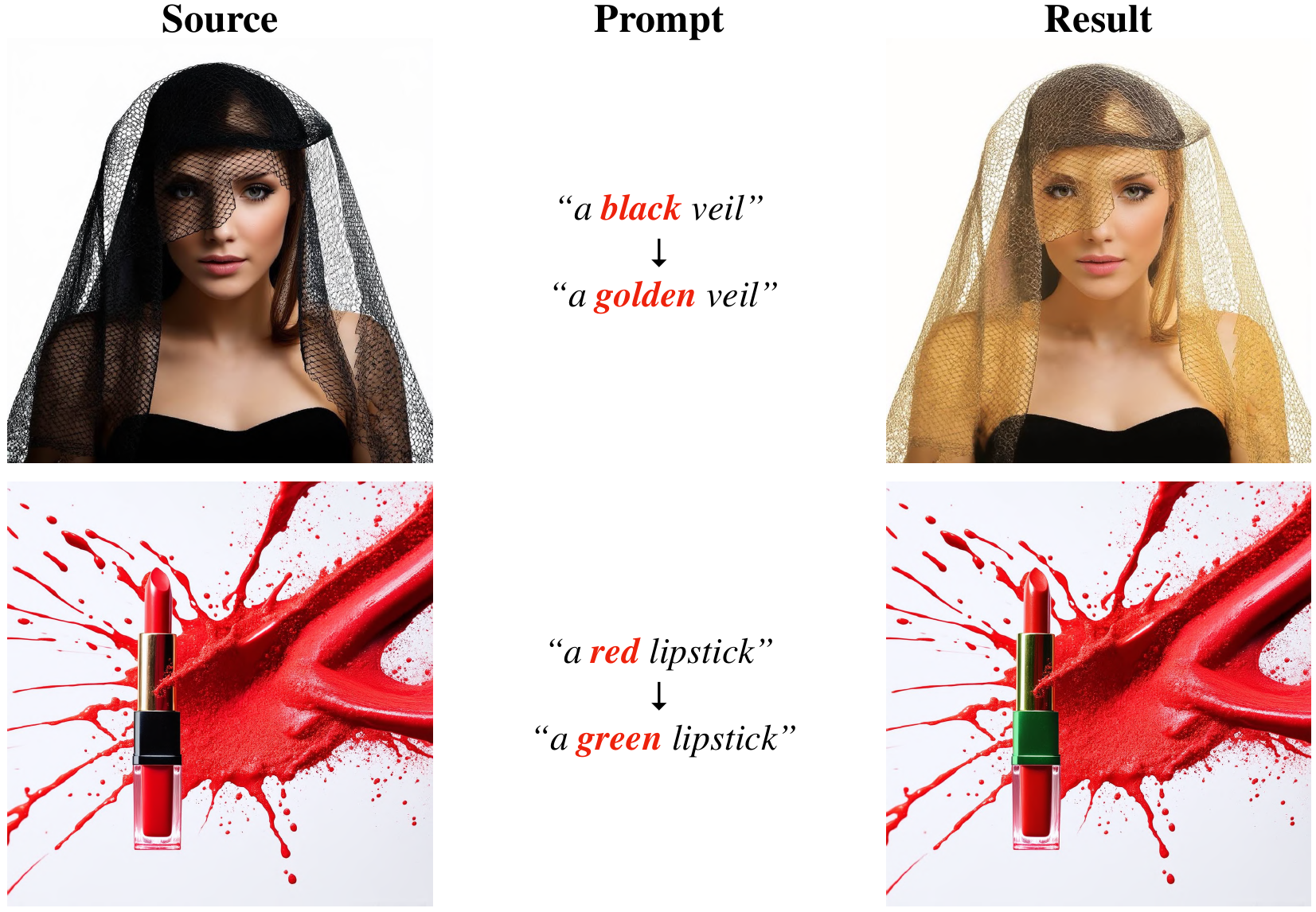}
  \caption{ Examples of typical failure cases. }
  \label{fig:limitation}
\end{figure} 

\subsection{Limitation}

The generation quality and the precision of text-guided localization in our method are ultimately constrained by the capabilities of the base generative models. Two representative failure modes are illustrated in Fig.~\ref{fig:limitation}:
\begin{itemize}[leftmargin=*]
\item \textbf{Localization Failure}: Small or abstract objects may not be edited when the attention map lacks a clear activation, leading to no visible change. For example, in the top case of Fig.~\ref{fig:limitation}, although the overall color, including some very small holes, is edited correctly, the model struggles to distinguish between intertwined hair and veil.
\item \textbf{Semantic Ambiguity}: Given a prompt to change lipstick color, the model may instead edit the lipstick case rather than the lipstick itself.
\end{itemize}
In a different aspect, compared with image models, current video models still lag considerably in generation fidelity. Nevertheless, as foundation models continue to improve, we expect our method to benefit correspondingly and expand its applicability.

Furthermore, our ability to edit real images and videos is inherently constrained by the limitations of current inversion and reconstruction techniques. Although our method performs reliably on data within the distribution of the generative model, editing real-world inputs requires accurately mapping them into the latent space of the model, a task that remains challenging and highly dependent on the quality of the inversion process.
\end{document}